\documentclass[lettersize,journal]{IEEEtran}
\usepackage{booktabs}
\usepackage{adjustbox}
\usepackage{hyperref}
\usepackage{amsmath,amsfonts}
\usepackage{algorithmic}
\usepackage{array}
\usepackage[caption=false,font=normalsize,labelfont=sf,textfont=sf]{subfig}
\usepackage{textcomp}
\usepackage{stfloats}
\usepackage{url}
\usepackage{verbatim}
\usepackage{graphicx}
\def\BibTeX{{\rm B\kern-.05em{\sc i\kern-.025em b}\kern-.08em
    T\kern-.1667em\lower.7ex\hbox{E}\kern-.125emX}}
\usepackage{balance}
\usepackage{orcidlink}

\def\MethodName{NeuralSRNF}
\newcommand{\s}{\ensuremath{\mathbb{S}}}

\newcommand{\stwo}{\ensuremath{\mathbb{S}^2}}
\newcommand{\sphere}{\stwo}
\newcommand{\sn}[1]{\s^{n}}
\newcommand{\domain}{\Omega}					% parameterization domain for these
\newcommand{\pointonsphere}{s}
\newcommand{\surface}{f} 	
\newcommand{\surfaces}{\Space{F}} % surface

\newcommand{\dinofeature}{\textbf{x}}

\newcommand{\srnf}{q}							% SRNF of a surface
\newcommand{\srnfone}{\srnf_1}							% SRNF of a surface
\newcommand{\srnftwo}{\srnf_2}							% SRNF of a surface
\newcommand{\srnfmap}{Q}                        			% SRNF map
\newcommand{\srnfsinverse}{\srnfmap^{-1}}                        			% SRNF map

\newcommand{\pointonsurface}{\surfacepoint} 
\newcommand{\surfaceone}{\surface_1} 						% surface
\newcommand{\surfacetwo}{\surface_2} 						% surface

\newcommand{\surfacepoint}{p}

\newcommand{\networkparams}{\Theta}
\newcommand{\srnfnetwork}{F_\networkparams}

\newcommand{\Euclidean}{\mathbf{L}^2}

\newcommand{\distance}{d}

\newcommand{\areameasure}{r}

\newcommand{\srnfs}{\mathcal{Q}}
\newcommand{\normalfield}{\textbf{n}}                   		 % Normal field on a surface

\newcommand{\shortpath}{\alpha}
\newcommand{\cov}{M}							% Covariance matrix
\newcommand{\eigenval}{\lambda}
\newcommand{\eigenvect}{\Lambda}					% eigenvector

\newcommand{\fundelm}{g}

\newcommand{\noi}{\noindent}

\newcommand{\eg}{\emph{e.g.,\ }}
\newcommand{\ie}{\emph{i.e.,\ }}

\newcommand{\etal}{\emph{et al.}}

\usepackage{framed, color}
\definecolor{shadecolor}{rgb}{1,0.5,0}

\newcommand{\Space}[1]{\ensuremath{\mathcal{#1}}}

\newcommand{\real}{\mathbb{R}}
\newcommand{\rplus}{\real_{>0}}

\newcommand{\rcubed}{\real^{3}}
\newcommand{\rthree}{\rcubed}

\newcommand{\ltwo}{\mathbb{L}^{2}}
\newcommand{\mean}{\mu}

\def\argmin{\mathop{\rm argmin}}

\begin{document}
\title{NeuralSRNF: Neural Square Root Normal Fields for the Statistical Shape Analysis and Generation of Nonrigid 3D and 4D Objects}
\author{%
  Awais Nizamani\,\orcidlink{0009-0005-0289-5930},
  Hamid Laga\,\orcidlink{0000-0002-4758-7510}, 
  Guanjin Wang\,\orcidlink{0000-0002-5258-0532},
  Farid Boussaid\,\orcidlink{0000-0001-7250-7407}, Senior Member, IEEE, \\
  Mohammed Bennamoun\,\orcidlink{0000-0002-6603-3257}, Senior Member, IEEE and
  Anuj Srivastava\,\orcidlink{0000-0001-7406-0338}, Member, IEEE%
  \thanks{
  	Awais Nizamani is with Murdoch University, Perth, Australia.
  	E-mail: Awais.Nizamani@murdoch.edu.au}
  \thanks{
  	Hamid Laga is with Murdoch University, Perth, Australia.
  	E-mail:    H.Laga@murdoch.edu.au.}

  \thanks{ Guanjin Wang is with Murdoch University, Perth, Australia.
  	E-mail: Guanjin.Wang@murdoch.edu.au.}

  \thanks{ Farid Boussaid is with The University of Western Australia, Perth, Australia.
  	E-mail: Farid.Boussaid@uwa.edu.au.}

  \thanks{ Mohammed Bennamoun is with The University of Western Australia, Perth, Australia.
  	E-mail: Mohammed.Bennamoun@uwa.edu.au.}

    \thanks{ Anuj Srivastava is with The Johns Hopkins University, Maryland, US.
  	E-mail: Anuj.Srivastava@jhu.edu.}
}

% \authorfooter{
%   %% insert punctuation at end of each item
%   \item
%   	Awais Nizamani is with Murdoch University.
%   	E-mail: Awais.Nizamani@murdoch.edu.au
%   \item
%   	Hamid Laga is with Murdoch University.
%   	E-mail:    H.Laga@murdoch.edu.au.

%   \item Guanjin Wang is with Murdoch University.
%   	E-mail: guanjin.wang@murdoch.edu.au.

%   \item Farid Boussaid is with The University of Western Australia.
%   	E-mail: farid.boussaid@uwa.edu.au.

%   \item Mohammed Bennamoun is with The University of Western Australia.
%   	E-mail: mohammed.bennamoun@uwa.edu.au.

%     \item Anuj Srivastava is with The Johns Hopkins University.
%   	E-mail: anuj.srivastava@jhu.edu.
% }

% \author{IEEE Publication Technology Department
% \thanks{Manuscript created October, 2020; This work was developed by the IEEE Publication Technology Department. This work is distributed under the \LaTeX \ Project Public License (LPPL) ( http://www.latex-project.org/ ) version 1.3. A copy of the LPPL, version 1.3, is included in the base \LaTeX \ documentation of all distributions of \LaTeX \ released 2003/12/01 or later. The opinions expressed here are entirely that of the author. No warranty is expressed or implied. User assumes all risk.}}

% \markboth{Journal of \LaTeX\ Class Files,~Vol.~18, No.~9, September~2020}%
% {How to Use the IEEEtran \LaTeX \ Templates}

\maketitle

\begin{abstract}
    We introduce \MethodName, a novel framework for the statistical shape analysis and generation of genus-zero 3D and 4D objects that undergo  nonrigid deformations. Traditional methods rely on complex and computationally expensive nonlinear elastic  metrics that  measure bending and stretching. Recent advances in elastic shape analysis achieve computational efficiency by mapping input 3D shapes to the space of Square Root Normal Fields (SRNFs) where the $\ltwo$ metric approximates the partial elastic metric, significantly facilitating the process of computing geodesics and summary statistics. SRNFs, however, are not invertible, and the numerical algorithms used to map SRNFs back to the original space of surfaces remain computationally very expensive and often lead to approximate results. This paper addresses this fundamental SRNF inversion problem using a novel neural representation, termed \MethodName. Unlike the commonly used numerical SRNF,   \MethodName~is  \textbf{(1)} continuous, and thus resolution-agnostic, enabling full functional shape analysis, \textbf{(2)} more accurate, and  \textbf{(3)} computationally more efficient as it can compute inverse SRNF maps along a geodesic path in less than $3$ s compared to over $10$ min for the numerical SRNF. We demonstrate, using various datasets, the utility and efficiency of the proposed \MethodName~in multiple elastic 3D and 4D shape analysis tasks such as geodesic computation, deformation transfer, statistical summaries computation, and 3D shape generation. We show that it outperforms competing methods on most evaluated datasets and metrics by a wide margin in both accuracy and computational efficiency. The source code and additional results are available at \url{https://awaisnizamani16.github.io/awais/NeuralSRNF/}.
\end{abstract}


\section{Introduction}
\label{sec:intro}

Shape is an important geometric property of natural and manmade objects. Understanding  differences and similarities between the shapes of objects has many applications in computer vision, augmented reality, biology, and anatomy. Shape analysis focuses on quantifying such differences or similarities,  and characterizing shape variability within a population of 3D and 4D objects using probability distributions. Traditional techniques achieve this by treating the shape of an object as a point on a Riemannian shape space equipped with a proper elastic metric that measures bending and stretching. This way, the dissimilarity between two shapes can be defined  as the optimal amount of bending and stretching that one needs to apply to one shape in order to align it with the other. State-of-the-art methods that follow this general pipeline exhibit two fundamental limitations. \textbf{First}, they treat 3D and 4D shapes as discrete surfaces and thus the metric is not invariant to the quality of the discretization. \textbf{Second}, the metric, which is nonlinear, is computationally very expensive to evaluate; thus, it becomes impractical for the statistical analysis of large shape datasets. 

Recent advances in elastic shape analysis achieve computational efficiency by mapping 3D shapes to the space of Square Root Normal Fields (SRNFs)~\cite{jermyn2012elastic,jermyn2017elastic,laga2017numerical} where  the $\ltwo$ metric approximates the partial elastic metric, significantly facilitating the process of computing
geodesics and summary statistics. SRNFs, however, are not invertible as there is no analytical expression for the inverse SRNF  for arbitrary points in $\ltwo$, and the injectivity and surjectivity of the SRNF map remain to be determined.  Thus, one cannot transfer geodesics and statistical analysis conducted in the SRNF space  back to the original space of surfaces. While  numerical algorithms have been used for the   SRNF  inversion~\cite{laga2017numerical}, they remain computationally very expensive and often lead to approximate results. %, thereby removing one of the main advantages of  SRNFs.  

This paper addresses this fundamental SRNF inversion problem using a novel neural representation, termed Neural Square Root Normal Fields (\MethodName).  Focusing on  genus-0 surfaces, which cover a wide range of shape classes, ranging from  human bodies to animals,  our key insight is to treat such surfaces as continuous functions parameterized using an MLP. This way, SRNFs become continuous functions. We then propose a novel network architecture that learns how to recover surfaces from their SRNF maps using a Conditional MLP. Unlike the commonly used numerical SRNF~\cite{laga2017numerical},   \MethodName~is  continuous. Thus, it is resolution-agnostic, enabling full functional shape analysis. The proposed approach is more accurate and computationally more efficient as it can  invert SRNF maps along a geodesic path in less than $3$ s compared to over $10$ min for the  numerical SRNF~\cite{laga2017numerical,nizamani2025DSNS}.

In summary, the  main contributions are as follows:
\begin{itemize}

    \item We treat genus-0 surfaces as continuous functions parameterized using MLPs. This provides a resolution-agnostic representation,  enabling full functional shape analysis.
    
    \item We propose a novel network architecture that provides a practical solution to the theoretical SRNF problem by learning to recover a valid surface from its SRNF representation. We achieve this using a conditional MLP  defined using DINOv2 encoded features and Feature-wise Linear Modulation (FiLM). Compared to~\cite{laga2017numerical}, ours is  more than $100$ times faster and more accurate. 
    
    \item We develop a  set of computational tools for  geodesic computation,  deformation transfer,  statistical summarization of 3D and 4D shape collections, and 3D shape generation.
\end{itemize}

\noi We demonstrate, using 3D and 4D human, face, and animal datasets, that  the proposed \MethodName~framework outperforms the state of the art by a wide margin in terms of accuracy and computational efficiency. 
% Please follow the steps outlined in this document very carefully when
% submitting your manuscript to Eurographics.

% You may as well use the \LaTeX\ source as a template to typeset your own
% paper. In this case we encourage you to also read the \LaTeX\ comments
% embedded in the document.

\begin{figure*}[t]
    \centering
    \includegraphics[width=0.99\linewidth,trim={0.5cm 12.5cm 0cm 3.5cm},clip]{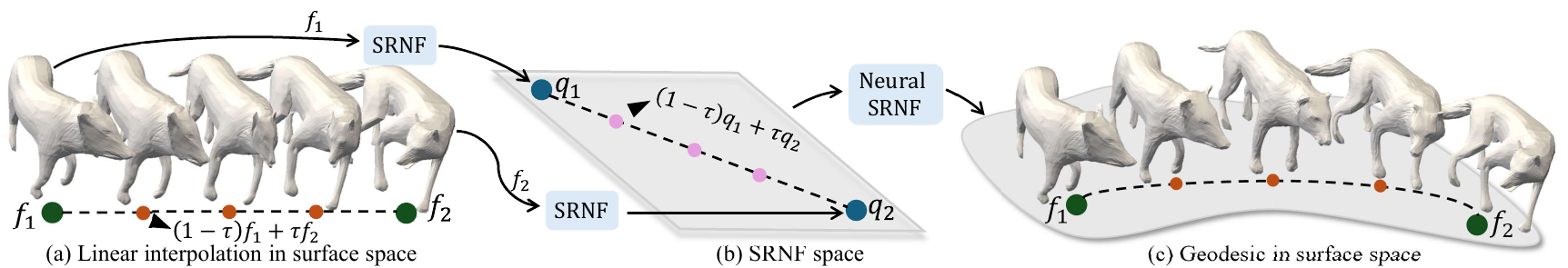}
    \caption{Illustration of the proposed \MethodName~framework. Given  surfaces $\surface_i$,  we first map them to the SRNF space using $\srnf_i = \srnfmap(\surface_i)$. We then compute geodesics between pairs $\srnfone$ and $\srnftwo$ in the SRNF space, which has an Euclidean structure,  and use \MethodName~to map their SRNFs back to the surface space. We show  \textbf{(a)} linear interpolation in the original space of surfaces,  \textbf{(b)} linear interpolation in the SRNF space, and \textbf{(c)} the geodesic computed by SRNF inversion of the linear path in the SRNF space.}% The surfaces are registered using $\gamma^*$, a registration framework, and $\tilde{\srnftwo}$ represents a spatially aligned SRNF.
     
    \label{fig:srnf_inversion_overview}
\end{figure*}

%-------------------------------------------------------------------------
\section{Related work}
\label{sec:related_work}

\noi\textbf{Statistical 3D shape analysis}   methods treat the shape of a 3D object as a point in a high-dimensional shape space equipped with a metric that measures physical deformations; see~\cite{laga2018survey} for a detailed review of the state-of-the-art.  Early approaches that are based on active shape models~\cite{COOTES199538} and morphable models~\cite{blanz1999morphable, egger20203dmorphable} assume that the shape space is Euclidean and use the $\ltwo$ metric to compute geodesics and  perform statistical analysis using Principal Component Analysis (PCA). These methods, however, are only suitable for rigid objects or objects that undergo small elastic deformations since the $\ltwo$ metric does not quantify elastic deformations. 
Later, more expressive metrics have been introduced to capture and quantify large nonrigid deformations. For example, Kilian \etal~\cite{kilian2007geometric} use a metric that penalizes deviations from rigidity and isometry. Drawing from the elasticity theory in physics, Wirth \etal~\cite{wirth2011continuum} and Zhang \etal~\cite{zhang2015shell} modeled stretching using the Cauchy-Green strain tensor and bending via differences in shape operators. Windheuser
\etal~\cite{windheuser2011geometrically} measured bending through variations in the mean curvature, while Heeren \etal~\cite{heeren2012time} used differences in the second fundamental form for bending and changes in the first fundamental form for stretching.

While being highly expressive, these physically motivated metrics are computationally intensive since geodesic computation and statistical summarization all require solving nonlinear optimization problems.  Jermyn \etal~\cite{jermyn2012elastic,jermyn2017elastic} introduced a more efficient approach by showing that the $\ltwo$ metric in the space of SRNFs is equivalent to a partial elastic metric defined as a weighted sum of bending and stretching. This significantly reduces the computational complexity: instead of working in the original space, one can map all the input shapes to the SRNF space, perform  the analysis tasks on that space using the  $\ltwo$ metric, and finally map the results back to the original space of shapes. The main theoretical limitation, however, is that this inverse mapping has no analytical expression, and its injectivity and surjectivity are yet to be established.  Laga \etal~\cite{laga2017numerical} addressed this  by using a numerical approach. The method, however, is computationally  expensive, and results in approximate solutions.  
This paper addresses this inversion problem using a learning approach, leading to a significantly faster and more accurate solution than the state-of-the-art. 

Note that statistical shape analysis methods require  spatially-registered surfaces. Some methods address this jointly with the statistical analysis~\cite{jermyn2012elastic,jermyn2017elastic,laga2017numerical}. Others assume that the registration is pre-established, \eg by using functional maps~\cite{Ovsjanikov2012Functional, Ovsjanikov2016Functional,DvirCyclicFunctionalMap, attaiki2023deepfunctionalmaps, magnet2023scalableefficientfunctionalmap, cao2023unsupervised, liu2025stablescore, cao2024spectralmeetsspatialharmonising}. 

\noi\textbf{Neural representations~\cite{park2019deepsdf,mildenhall2020nerf,wang2021neus}} leverage neural networks to represent the geometry and appearance of 3D shapes in a resolution-agnostic manner. While they were primarily intended for 3D and 4D  reconstruction, and novel view synthesis, some recent papers have begun to explore their potential in statistical shape analysis. In particular, Williamson \etal~\cite{williamson2024sns} showed how
to compute differential properties of neural surfaces and perform
geometry processing tasks on spherical neural representations without upfront discretization. Sang \etal~\cite{sang2025implicit,sang20254deform} used neural implicit functions to learn how to deform a source 3D surface to a target one. Their method, however, requires training for every pair of 3D objects and is therefore computationally very expensive. Nizamani \etal~\cite{nizamani2025DSNS} introduced dynamic neural surface  representations for the statistical analysis of 4D shapes. Some recent approaches embed surfaces into a learned latent space and compute geodesics within latent space using ARAP regularizations~\cite{eisenberger2021neuromorph, cosmo2020LIMP, Yang2023GeoLatent}, Laplace-Beltrami operators~\cite{lemeunier2022SAE}, and path straightening~\cite{hartwig2026geodesiccalculusimplicitlydefined}. These geometric constraints are also computationally expensive. 

This paper uses the surface-based neural representation of Nizamani \etal~\cite{nizamani2025DSNS} but focuses on addressing the SRNF inversion problem using a novel neural representation termed \MethodName. This leads to more accurate results while being significantly faster (up to $100$ times) compared with the state-of-the-art methods.

%-------------------------------------------------------------------------
\section{Method}

We treat the shape of a closed genus-0 surface $\surface: \domain =\stwo \to \rthree$, after normalization for translation and scaling,  as an element of a pre-shape space $\surfaces$. To measure the dissimilarity between two surfaces $\surfaceone$ and $\surfacetwo$ in $\surfaces$, and subsequently compute geodesic paths, we equip $\surfaces$ with a Riemannian elastic metric that quantifies the amount of bending and stretching one needs to apply to $\surfaceone$ in order to align it onto $\surfacetwo$. This approach, however, is computationally expensive, making it hard to use in downstream tasks such as the computation of geodesics and statistical summaries. This computational cost stems from two fundamental limitations. \textbf{First}, the computational complexity of the metric scales exponentially with the resolution of the discretization of the domain $\domain$. \textbf{Second}, the metric is highly nonlinear. Thus, computing geodesics and summary statistics requires solving complex nonlinear optimizations.
To address the computational complexity issue, 
Jermyn \etal~\cite{jermyn2012elastic}  map 3D shapes to the space $\srnfs$ of SRNFs where the $\ltwo$ metric in $\srnfs$  is equivalent to the partial elastic metric in $\surfaces$. This significantly simplifies downstream tasks since these can be performed in the SRNF space, using the  $\ltwo$ metric, and then mapped  back to the space of surfaces for visualization. The fundamental problem, however, is that SRNFs are not invertible, \ie the mapping from the SRNF space to the space of surfaces has no analytical form.  Laga \etal~\cite{laga2017numerical} approximate this inversion using a numerical approach but it is very slow and often leads to inaccurate results. 

This paper jointly addresses these two problems by introducing a novel neural representation termed \MethodName, which represents surfaces  $\surface \in \surfaces$ in a resolution-agnostic manner, enabling full functional shape analysis, and learns how to invert SRNF maps in a computationally efficient manner while achieving significantly better accuracy than numerical methods (Section~\ref{sec:srnf_inversion}).  We will  show how to use the proposed \MethodName~in computing geodesics (Section~\ref{sec:geodesics}), deformation transfer (Section~\ref{sec:deformation_transfer}), and summary statistics and shape generation (Section~\ref{sec:summary_statistics}).

\begin{figure*}[t]
    \centering
    % \includegraphics[width=0.95\linewidth, trim=0cm 12.5cm 1.7cm 2.5cm, clip]{figures/neural_srnf_architecture.pdf}
    % \caption{Illustration of the \MethodName~network architecture. \textbf{(a)} \MethodName~recovers the neural surface $\surface$ for a given SRNF map $\srnf$. \textbf{(b)} The detailed network architecture of \MethodName. Given a point $\pointonsphere$, its corresponding SRNF value $\srnf(\pointonsphere)$, and DinoV2 encoded latent code $\dinofeature$, we map it to the point on the surface $\pointonsurface$. \textbf{(c)} A close-up view of the Feature-wise Linear Modulation (FiLM) layer.}

    \includegraphics[width=\linewidth,trim={3.8cm 13cm 2cm 2cm},clip]{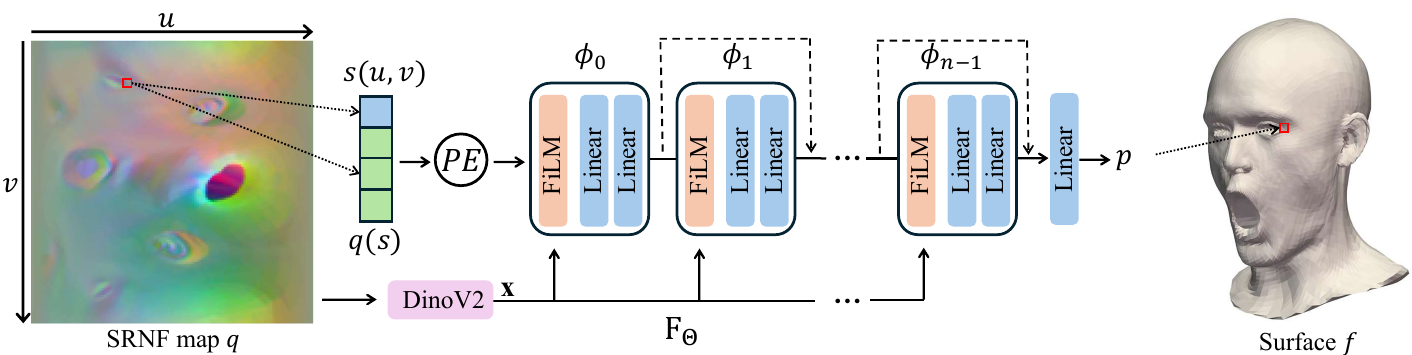}
    \caption{Network architecture of \MethodName, which takes an SRNF map $\srnf$ and recovers a surface $\surface$  whose SRNF map is as close as possible to $\srnf$.}
    
    \label{fig:neural_srnf_net}
\end{figure*}

\subsection{The SRNF inversion problem}
\label{sec:srnf_inversion}

Given a surface $\surface\in \surfaces$, its SRNF $\srnf = \srnfmap(\surface)$ is essentially the normal field on the surface scaled by the square root of the norm of the normals. In other words, 
\begin{equation}
    \srnfmap: \surfaces \to \srnfs  \text{ such that }\srnfmap(\surface)(\pointonsphere) = \frac{\normalfield(\pointonsphere)}{ \parallel\normalfield(\pointonsphere)\parallel^{\frac{1}{2}}}.
\end{equation}

\noi Here, $\srnfs$ is the space of SRNFs, $\normalfield = \frac{\partial \surface}{\partial u} \times \frac{\partial f}{\partial v}$ is the normal vector field to the surface,   $\hat\normalfield$ is the unit normal field, and 
$\pointonsphere =  (u, v) \in \stwo$. 
Note that the SRNF is physically motivated since \textbf{(1)} changes in the direction of the normal vector $\normalfield(\pointonsphere)$ at a given point $\pointonsphere$ on the surface induce local bending, while \textbf{(2)} changes in the  norm of a normal vector, denoted by $\areameasure(\pointonsphere) = \|\normalfield(\pointonsphere) \|$, which is the square root of the determinant of the first fundamental form $\fundelm$ of the surface at $\pointonsphere$, result in local stretching. Jermyn \etal~\cite{jermyn2012elastic} showed that the partial elastic metric $ \distance_{\surfaces}$ that quantifies the weighted sum of bending and stretching between two surfaces $\surfaceone$ and $\surfacetwo$ in $\surfaces$ is equivalent to the $\ltwo$ metric in the space $\srnfs$ of SRNFs, \ie,
\begin{equation}
    \distance_{\surfaces}(\surfaceone, \surfacetwo) \equiv\parallel \srnfone - \srnftwo \parallel, \text{ where } \srnf_i = \srnfmap(\surface_i).
\end{equation}

\noi With this formulation, the geodesic, or the shortest path with respect to the metric $\distance_{\surfaces}$, between $\surfaceone$ and $\surfacetwo$ is equivalent to the straight line between $\srnfone$ and $\srnftwo$. Thus, working in the SRNF space is computationally very attractive. However, the following questions about the SRNF inversion problem, \ie how to map results from the SRNF space back to the original space of surfaces, remain unaddressed: Given an SRNF $\srnf$, or equivalently a unit normal field $\hat{\normalfield}: \sphere \to \sphere$ and a local area element $\areameasure = \parallel\normalfield\parallel: \sphere \to \rplus$, 
\begin{enumerate}
    \item Under what conditions does the pair $(\hat{\normalfield}, \areameasure)$ correspond to a valid surface $\surface: \sphere \to \rthree$?

    \item Does the differential equation $\normalfield = |\srnf| \srnf$ admit a solution, and is it unique up to translation?
\end{enumerate}

\noi In the \textbf{first} case, if we are given $(\hat{\normalfield},\fundelm)$, it is known that the representation is injective (up to isometries of $\rthree$), \ie the original surface can be reconstructed up to translation and rotation~\cite{arnold2004problems}. This is not the case with the SRNF, or equivalently $(\hat{\normalfield}, \areameasure)$, where $\areameasure$ is unable to fully measure the stretching. 
The \textbf{second} case considers the existence of a unique global solution for the inverse of the SRNF $\srnf$, which clearly does not always exist. For a small patch $\srnf(\pointonsphere)$, one can reconstruct the surface $\surface(\pointonsphere)$, but it may not be unique. Also, this is not sufficient for shape analysis as one requires a global solution for an entire surface.

\begin{figure}[t]
    \centering
    \includegraphics[width=0.95\linewidth,trim={0cm 5.25cm 8.25cm 2.5cm},clip ]{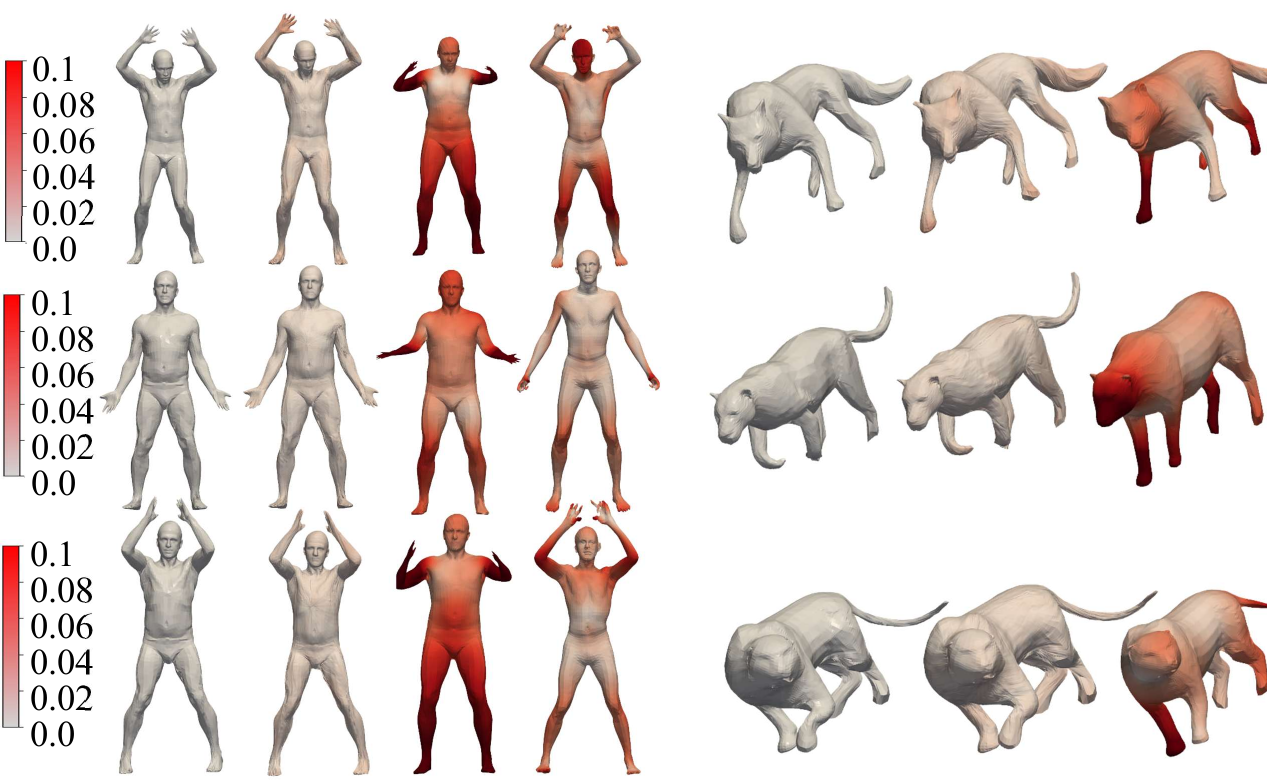}
    % \begin{adjustbox}{width=0.9\textwidth}
        \begin{tabular}{@{\hspace{1.2cm}} l @{\hspace{0.5cm}} c @{\hspace{0.5cm}} c @{\hspace{0.5cm}} c @{\hspace{1.1cm}} l @{\hspace{0.9cm}} c @{\hspace{0.8cm}} c  @{\hspace{6cm}} }
         (a)   & (b) &  (c) & (d)  &  (a)  & (b) & (c) \\
        \end{tabular}    
    % \end{adjustbox}
    
    \caption{Accuracy of the SRNF inversion on 3D humans and animals. \textbf{(a)} Ground-truth surfaces.  \textbf{(b)}  Reconstructed surfaces using  our \MethodName~method. \textbf{(c)} Reconstructed surfaces using the numerical SRNF~\cite{laga2017numerical}. \textbf{(d)} Reconstructed surfaces using the spectral autoencoder~\cite{lemeunier2022SAE}. The pointwise error is shown as a heatmap for (b), (c), and (d).}
    
    % \textbf{(a)} The ground truth surfaces. \textbf{(b)} Reconstructed surfaces using the numerical SRNF~\cite{laga2017numerical}.  \textbf{(c)}  Reconstructed surfaces using  our \MethodName~method. The pointwise error is shown as a heatmap for (b) and (c). The Supplementary Material provides additional results.}
    \label{fig:neural_srnf_quality_mainl}
\end{figure}

\begin{table*}
    \centering
    \resizebox{\textwidth}{!}{
    \begin{tabular}{@{}lc|cccc|cccc@{}}
    \toprule
    \textbf{Dataset} &  \textbf{Spherical} &
    \multicolumn{4}{c|}{\textbf{Numerical SRNF~\cite{laga2017numerical}}} & 
    \multicolumn{4}{c}{\textbf{\MethodName\ (Ours)}} \\
    % \cline{2-9}
     & \textbf{resolution}& \textbf{Mean} & \textbf{Std} & \textbf{Median } & \textbf{Time (sec)} 
     & \textbf{Mean}  & \textbf{Std } & \textbf{Median} & \textbf{Time (sec)} \\
    \hline
    COMA~\cite{COMA:ECCV18}                 & $128\times128$& $0.058 \times 10^{-3}$     & $0.057 \times 10^{-3}$   & $0.041 \times 10^{-3}$       & $4.73$  & $\mathbf{0.002  \times 10^{-3}}$   &  $\mathbf{0.001 \times 10^{-3}}$   & $\mathbf{0.002  \times 10^{-3}} $    &    $\mathbf{0.10}  $\\
    CAPE~\cite{CAPE:CVPR:20}                & $512\times512$& $5.433 \times 10^{-3}$     & $5.608 \times 10^{-3}$   & $3.275 \times 10^{-3}$       & $67.22$ & $\mathbf{0.336   \times 10^{-3}}$  &  $\mathbf{0.361 \times 10^{-3}}$  & $\mathbf{0.242   \times 10^{-3}} $    &    $\mathbf{0.65}   $\\
    DFAUST~\cite{dfaust:CVPR:2017}          & $512\times512$& $2.655 \times 10^{-3}$     & $1.885 \times 10^{-3}$   & $2.194  \times 10^{-3}$      & $69.10$ & $\mathbf{0.043   \times 10^{-3}}$  &  $\mathbf{0.191 \times 10^{-3}}$  & $\mathbf{0.036  \times 10^{-3}}  $    &    $\mathbf{0.60}    $\\
    Animal (w tail)~\cite{xu2023animal3d}   & $128\times128$& $2.728 \times 10^{-3}$     & $1.861 \times 10^{-3}$   & $2.350  \times 10^{-3}$      & $15.93$ & $\mathbf{0.078   \times 10^{-3}}$ &  $\mathbf{0.112 \times 10^{-3}}$   & $\mathbf{0.042 \times 10^{-3}}  $    &    $\mathbf{0.11}   $\\
    Animal (w/o tail)~\cite{xu2023animal3d} & $128\times128$& $1.208 \times 10^{-3}$     & $1.192 \times 10^{-3}$   & $0.812  \times 10^{-3}$      & $12.77$ & $\mathbf{0.030   \times 10^{-3}}$ &  $\mathbf{0.074 \times 10^{-3}}$  & $\mathbf{0.012  \times 10^{-3}} $     &    $\mathbf{0.11}    $\\
    \bottomrule
    \end{tabular}
    }
    \caption{Comparison of the proposed \MethodName\ inversion with numerical SRNF~\cite{laga2017numerical}. The evaluation is performed on $100$ test samples from five datasets. We measure the mean squared $\Euclidean$ distance and report the mean, standard deviation, and median inversion errors, together with the average computation time for both methods.  }
    % average time taken as well as the mean, median, and standard deviation of the inversion error for both methods. %Note that due to the fact that the numbers are small, the mean, std, and median values have been scaled by a factor of  $(\times 10^{3})$.
   
    \label{tab:comparison}
\end{table*}

\subsubsection{Neural Square Root Normal Fields}
\label{sec:neural_srnf}

In the absence of a theoretical solution, we reformulate the SRNF inversion problem as an optimization problem, \ie  given an SRNF $\srnf \in \srnfs$, we seek to find a surface $\surface^* \in \surfaces$ whose SRNF is as close as possible to $\srnf$:
\begin{equation}
    \surface^* = \argmin_{\surface \in \surfaces} \| q - \srnfmap(\surface) \|^2.
    \label{eq:srnf_inv_optimization}
\end{equation}

\noi When dealing with non-degenerate closed surfaces, Laga \etal~\cite{laga2017numerical} solved this optimization using a numerical approach. However, the approach is slow and requires a multiresolution and multiscale representation of the surfaces, in addition to a careful initialization of the optimization procedure. These significantly limit its applicability.

In this paper, we adopt a different approach. Instead of seeking to solve the optimization problem of Eqn.~\eqref{eq:srnf_inv_optimization}, we propose to explicitly learn the inverse mapping that maps the subset of $\srnfs$ that corresponds to valid surfaces to their corresponding surfaces in $\surfaces$. We  treat this inverse mapping as a set of functions $\srnfnetwork$, parameterized by the set of parameters $\networkparams$, of the form:
$
    \srnfnetwork: ( \pointonsphere,\srnf(\pointonsphere); \dinofeature) \to \surface(\pointonsphere)$.
They take a point $\pointonsphere \in \stwo$ and its corresponding SRNF $\srnf(\pointonsphere)$, and find its corresponding point $\surface(\pointonsphere)$ on the surface. By conditioning $\srnfnetwork$ on the shape-specific latent code $\dinofeature$, one can represent many shapes with a single function.

\begin{figure}
    \centering
    \includegraphics[width=0.95\linewidth,trim={0.25cm 3cm 0cm 2cm},clip ]{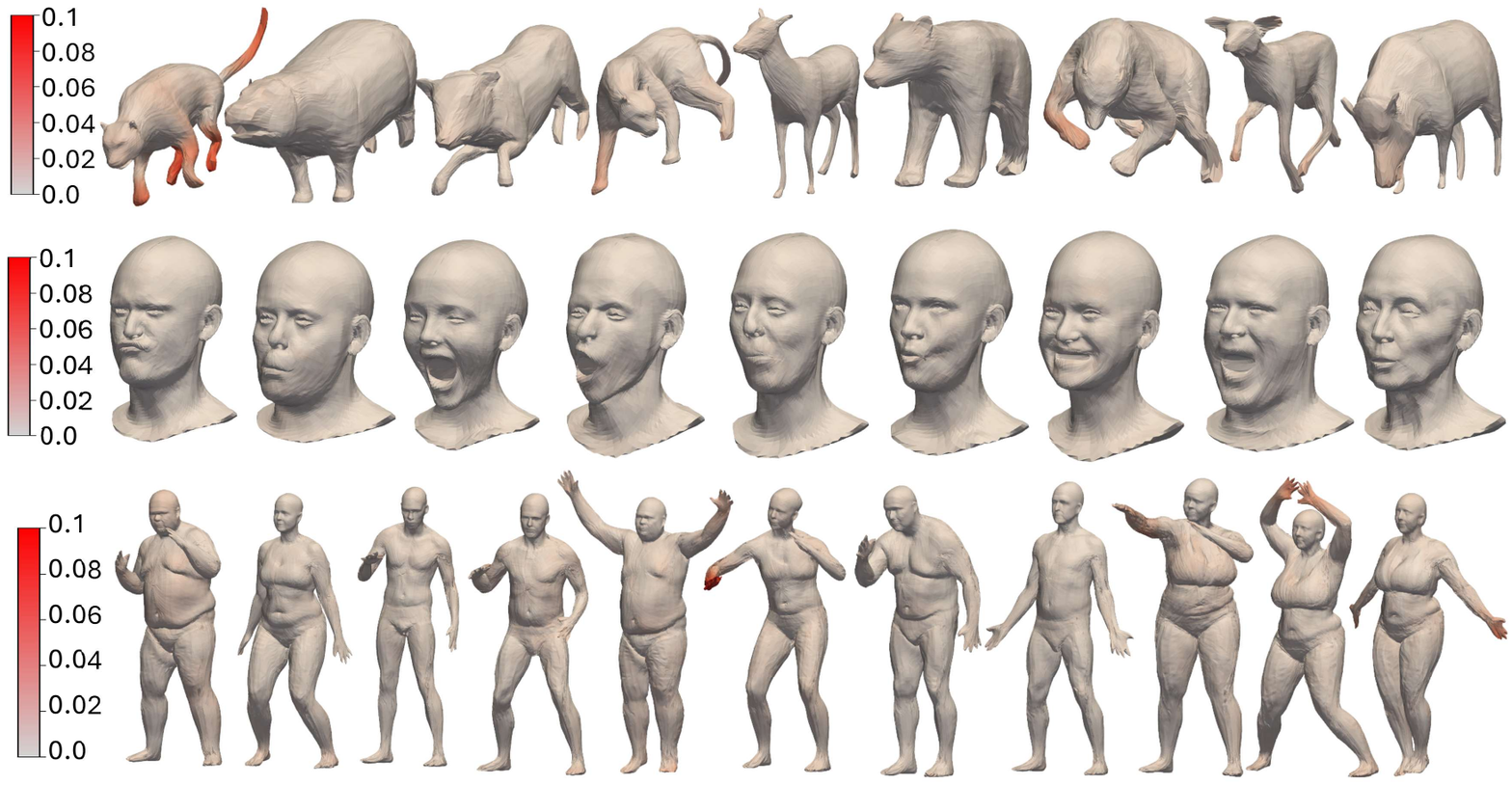}
    \caption{We show the qualitative results of the proposed \MethodName~inversion on animals, faces, and humans. The error is shown as a heatmap computed as the pointwise distance between the ground-truth and reconstructed surface using our \MethodName~method.}
    \label{fig:neural_srnf_quality_suppl}
\end{figure}

Formally, we represent $\srnfnetwork$  using an MLP, conditioned on the latent code $\dinofeature$, 
and learn its parameters $\networkparams$ from input SRNFs $\{ \srnf_1, \dots ,\srnf_n\}$ paired with their corresponding ground-truth surfaces $\{ \surface_1, \dots ,\surface_n\}$.
The MLP is composed of eight residual blocks, each containing two fully connected layers with $1024$ hidden units,  and a FiLM conditioning layer~\cite{perez2018film} to  allow the network to adapt its intermediate representations in each residual block, similar to the feature modulation strategy used in pi-GAN~\cite{chanmonteiro2020pi-GAN}.  Instead of using the ReLU activation function, which results in sharp edges, we employ the SoftPlus activation~\cite{zheng2015softplus} to ensure a smooth, continuous, and differentiable surface representation. Lastly, a linear projection layer maps the network output to a 3D point. Figure~\ref{fig:neural_srnf_net} summarizes this  architecture.

Inspired by NeRF~\cite{mildenhall2020nerf}, we apply  positional encoding to the SRNF point $\srnf(\pointonsphere)$ and to the coordinates $\pointonsphere$. To enable the network to invert any SRNF map, we condition it on the shape latent code $\dinofeature$. Unlike  DeepSDF~\cite{park2019deepsdf}, which learns the latent code in an auto-decoding fashion,  our latent codes are pre-computed from the SRNF map of each shape using a pretrained DINOv2~\cite{oquab2023dinov2} encoder. We found that this approach leads to faster computation time as the latent codes do not require finetuning during inference. The Supplementary Material provides a detailed ablation study of the network architecture, including the role of FiLM-based conditioning and the importance of DINOv2 features as latent codes.

\subsubsection{Training and losses}

Let $\{ \srnf_1, \dots ,\srnf_n\}$ be a set of SRNFs,   $\{ \surface_1, \dots ,\surface_n\}$  their corresponding spherically-parameterized ground-truth surfaces, and $B$ the batch size.  Each SRNF $\srnf_j$ is first encoded using DINOv2 encoder to obtain a  latent code $\dinofeature_j$. We follow the DINOv2 input format and  normalize  $\srnf_j$ to keep it within the range of $[0,1]$ (the normalization is only applied to generate the latent code $\dinofeature_j$). To create a continuous representation, we sample $K$ spherical coordinates $\{\pointonsphere_{i} \in \stwo\}_{i=1}^K$. 
For each spherical coordinate $\pointonsphere_{i} $, we obtain its corresponding points $\pointonsurface_{ij} = \surface_j(\pointonsphere_i)$ on the input surfaces, and their respective SRNF values $\srnf_j(\pointonsphere_i)$.
Next, we train  the MLP  $ \srnfnetwork$ by minimizing the mean squared error (MSE) between the ground-truth point $\pointonsurface_{ij}$ and the predicted point $\pointonsurface_{ij}^{*}$:

\begin{equation}
    L_{\text{MSE}} = \frac{1}{B}\frac{1}{K}\sum_{j=1}^B \sum_{i=1}^K \parallel \srnfnetwork( \pointonsphere_{i},\srnf_{j}(\pointonsphere_{i}), \dinofeature_j) - 
  p_{ij} \parallel^2. 
\end{equation}

\noi One advantage of the \MethodName\ is that it is a continuous and resolution-agnostic function. Thus, one can compute  all its  differential properties, including the surface normals,  using automatic differentiation. 

\subsection{Geodesics}
\label{sec:geodesics}

We define a path in $\surfaces$ between two surfaces $\surfaceone$ and $\surfacetwo$   as a curve $\shortpath_{\surfaceone \to \surfacetwo} : [0,1] \to \surfaces$ such that $\shortpath(0) = \surfaceone$, $\shortpath(1) = \surfacetwo$, and $\shortpath(\tau) \in \surfaces$ for all  $\tau \in [0,1]$. The geodesic between the two surfaces, \ie the minimum amount of bending and stretching one needs to apply to $\surfaceone$ in order to align it onto $\surfacetwo$, is then the shortest path under the elastic metric between the two surfaces. However, finding this shortest path using algorithms such as path straightening requires solving a highly nonlinear optimization problem since the metric is nonlinear. 
To address this problem in a computationally efficient manner, we first map the two surfaces to the SRNF space, resulting in $\srnfone = \srnfmap(\surfaceone)$ and $\srnftwo = \srnfmap(\surfacetwo)$. Since the $\ltwo$ metric in $\srnfs$ is equivalent to the elastic metric in the original space $\surfaces$ of surfaces, one only needs to compute the straight line $\shortpath_{\srnf}$ between $\srnfone$ and $\srnftwo$:
\begin{equation}
    \shortpath_{\srnf}(\tau) = (1 - \tau)\srnfone + \tau \srnftwo, \quad \tau \in [0,1].
    \label{equ:srnf_path}
\end{equation}

\noi We then map  $\shortpath_{\srnf}(\tau) \in \srnfs$ for all $\tau \in [0, 1]$ back to the original space $\surfaces$ of surfaces using the proposed \MethodName, \ie $\shortpath_{\surface}(\tau) = \srnfsinverse(\shortpath_{\srnf}(\tau))$. Figure~\ref{fig:srnf_inversion_overview}-(c) shows an example of such a geodesic path between $\surfaceone$ and $\surfacetwo$.

\subsection{Deformation transfer}
\label{sec:deformation_transfer}

The proposed framework can  be used to transfer deformations across surfaces. This problem can be defined as follows: we are given a source deformation $\surfaceone \to h_1$, \ie a surface $\surfaceone$ that deforms into the surface  $h_1$, and  we would like to deform a surface $\surfacetwo$ so that it assumes the same pose as $h_1$, obtaining another surface $h_2$. Traditionally, this is solved using parallel transport, \ie we first compute the geodesic between  $\surfaceone$ and $h_1$, and parallel transport it to $\surfacetwo$. This involves computing geodesics in $\surfaces$ and parallel-transporting them, thus it  is computationally expensive.
With the proposed \MethodName, this parallel transport problem can be solved in a very efficient manner. We first map the surfaces to the SRNF space $\srnfs$, which has an $\ltwo$ structure. Then, the SRNF of $h_2$ is given by:
\begin{equation}
    \srnfmap(h_2) = \srnfmap(\surfacetwo) + (\srnfmap(h_1) - \srnfmap(\surfaceone)).
\label{equ:defomation_transfer}
\end{equation}

\noi Here, $(\srnfmap(h_1) - \srnfmap(\surfaceone))$ represents the deformation field $v$, which is added to $\srnfmap(\surfacetwo)$ to obtain the deformed target surface $\srnfmap(h_2)$.  $\srnfmap(h_2)$ can then be inverted back to the space of surfaces  using the proposed \MethodName\ inversion. % $\srnfsinverse(\srnfmap(h_2))$.

Note also that one can define a path $\shortpath_{\srnf}(\tau) = \srnfmap(\surfacetwo) + \tau(\srnfmap(h_1) - \srnfmap(\surfaceone)) $. When $\tau \in [0, 1]$, it is equivalent to transporting the entire geodesic between $\surfaceone$ and $h_1$ to $\surfacetwo$. For $\tau > 1.0 $, this is equivalent to extrapolating the deformation beyond the given one.

\begin{figure}
    \centering
    \includegraphics[width=0.7\linewidth,trim={0cm 4cm 0cm 2cm},clip ]{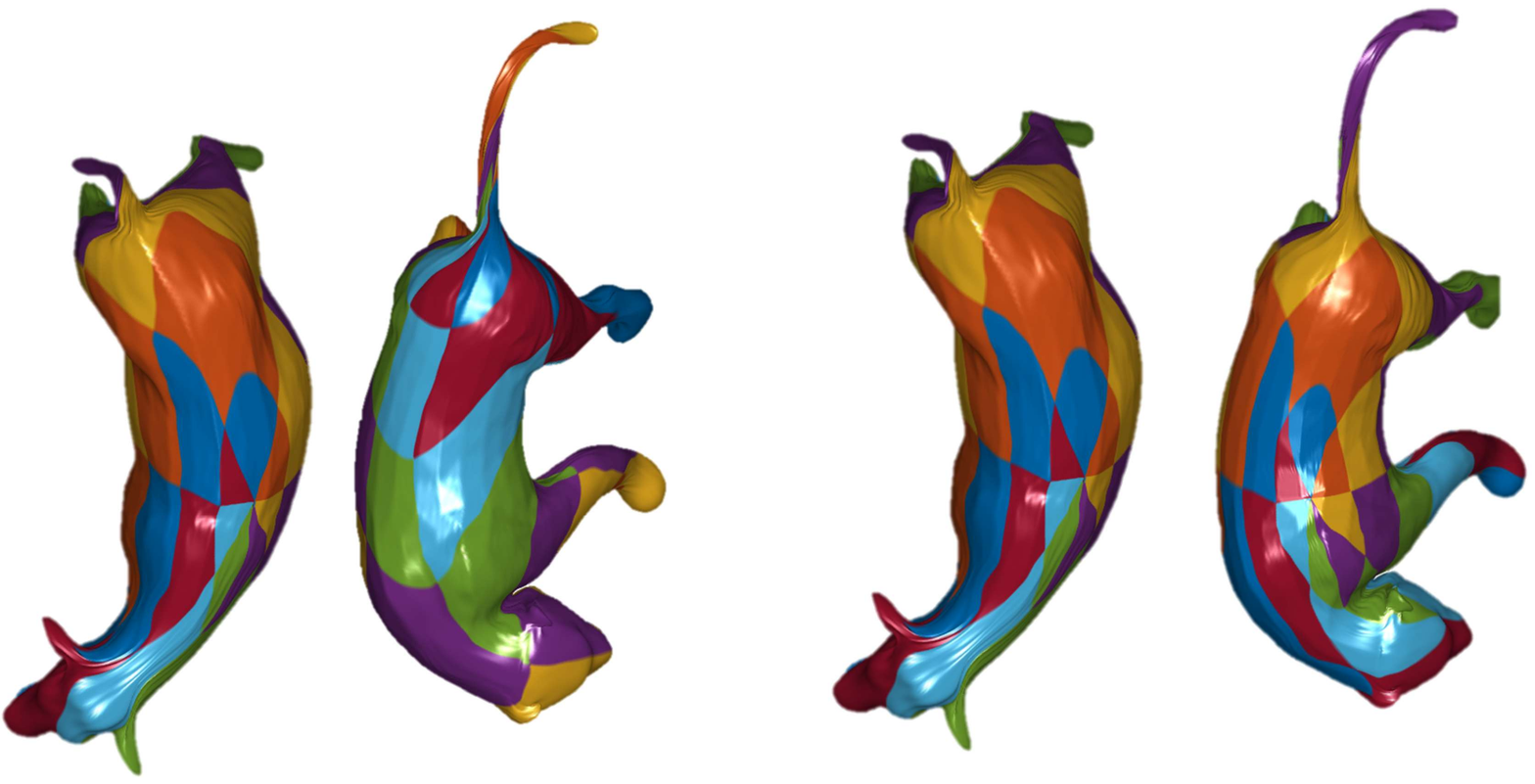}
    \begin{tabular*}{\linewidth}{@{}p{0.5\linewidth} p{0.35\linewidth}@{}}
        \centering  \textbf{(a)} Before registration. &
        \centering  \textbf{(b)} After registration. 
    \end{tabular*}\\
    \includegraphics[width=0.8\linewidth,trim={0cm 7.5cm 4cm 2cm},clip ]{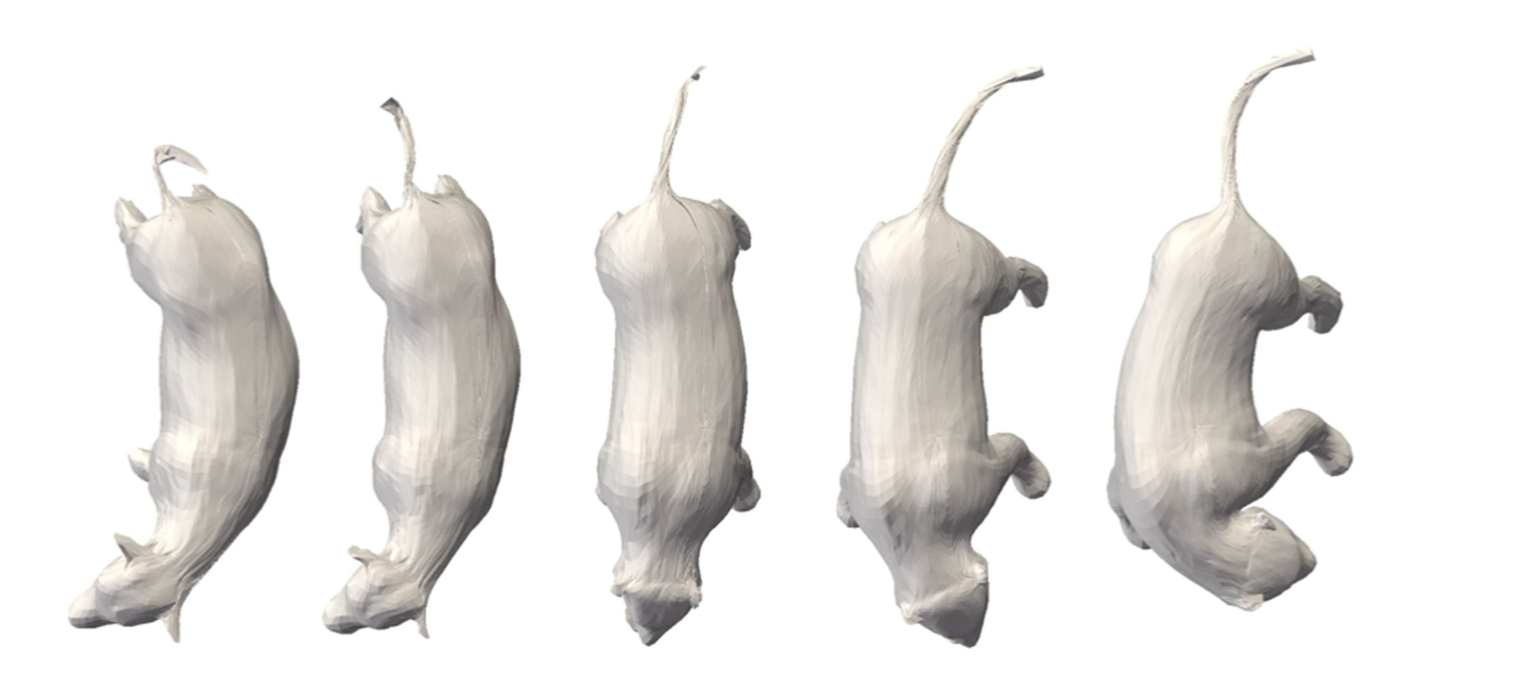}\\
    \centering \textbf{(c)} Geodesic after registration.
    \caption{Joint registration and geodesic computation in the SRNF space.}
    
    \label{fig:animal_geodesic_with_correspondence}
\end{figure}

\begin{figure}
    \centering
    \includegraphics[width=\linewidth,trim={0cm 0cm 0cm 0cm},clip ]{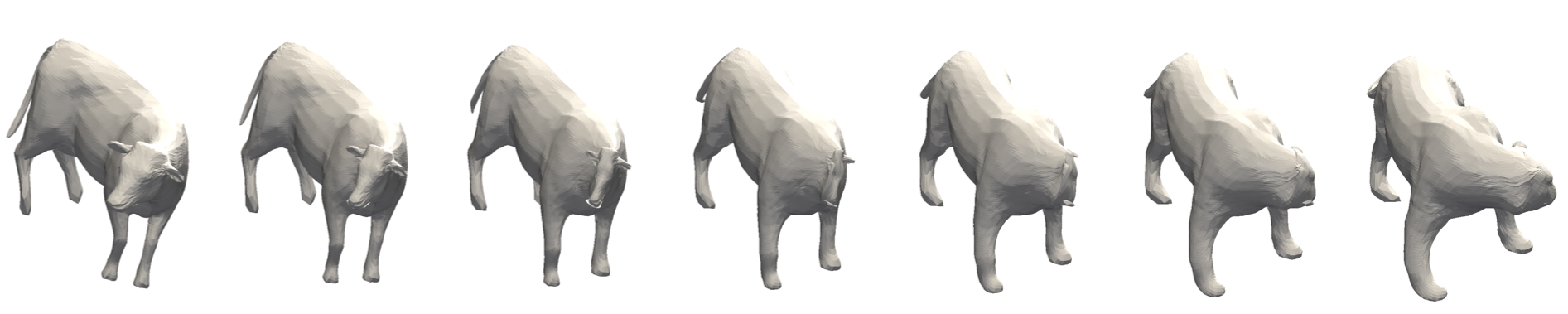}\\
    \textbf{(a)} Linear path.\\
    \includegraphics[width=\linewidth,trim={0cm 0cm 0cm 0cm},clip ]{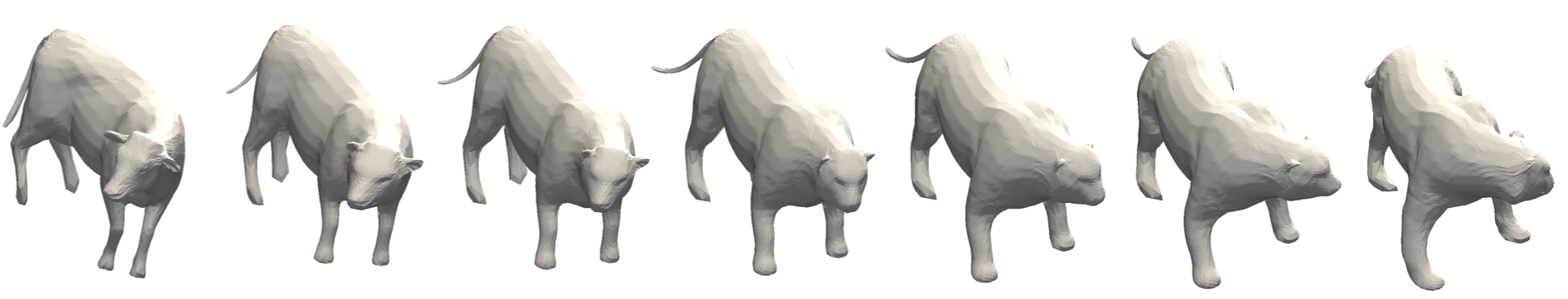}\\
    \textbf{(b)} Geodesic using numerical SRNF~\cite{laga2017numerical} ($118$ s).\\
    
    \includegraphics[width=\linewidth,trim={0cm 0cm 0cm 0cm},clip ]{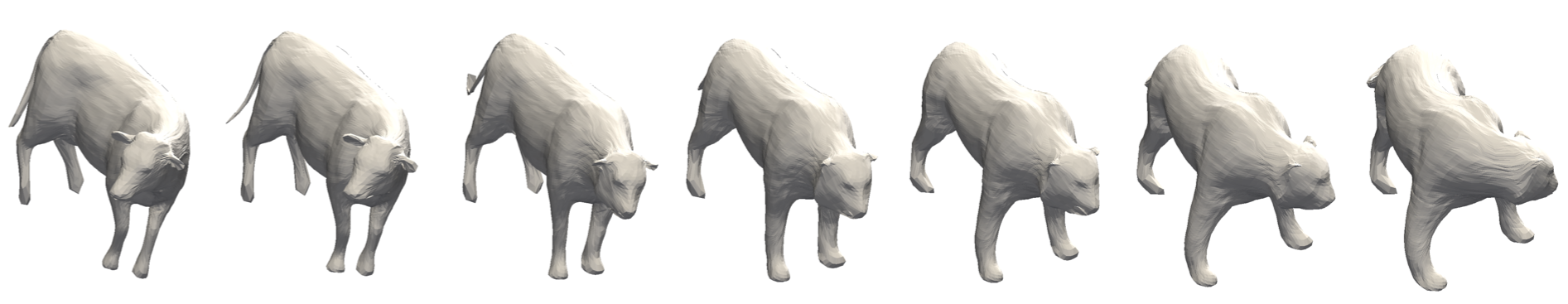}\\
   \textbf{(c)} Geodesic using our \MethodName~($2$ s).
    \caption{Examples of geodesics generated with our method and the state of the art. Observe that the intermediate surfaces of our approach are more natural.}
    % with a $50$ times faster speedup; see the  Supplementary Material for additional results.}
    \label{fig:animal_1_geodesics_main}
\end{figure}

\begin{figure}
    \centering
    % \includegraphics[width=0.95\linewidth,trim={0 0 0 0},clip ]{figures/Suppl/geodesics/animal/2/linear.png}\\
    % \textbf{(a)} Linear path.
    \includegraphics[width=0.95\linewidth,trim={0 0 0 0 },clip ]{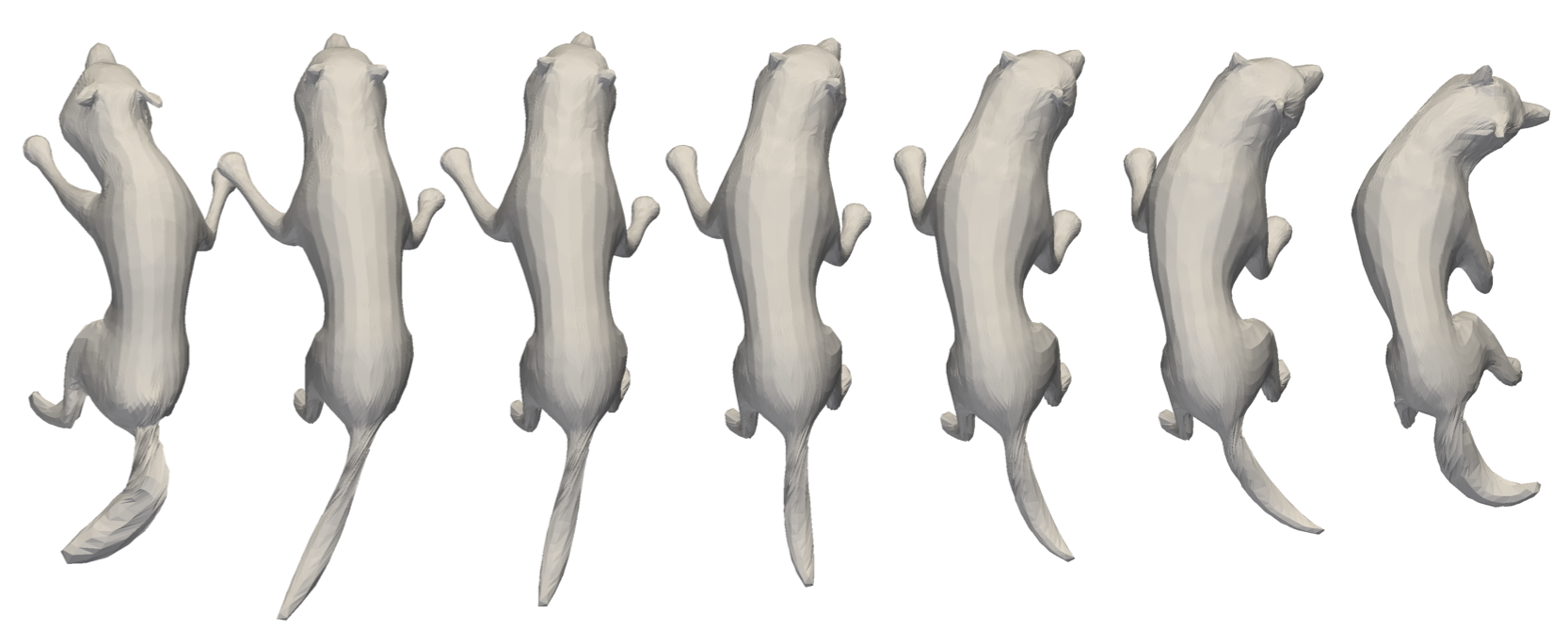}\\
    \textbf{(a)} Geodesic using numerical method~\cite{laga2017numerical} ($118$ s).
    \includegraphics[width=0.97\linewidth,trim={0 0 0 0},clip ]{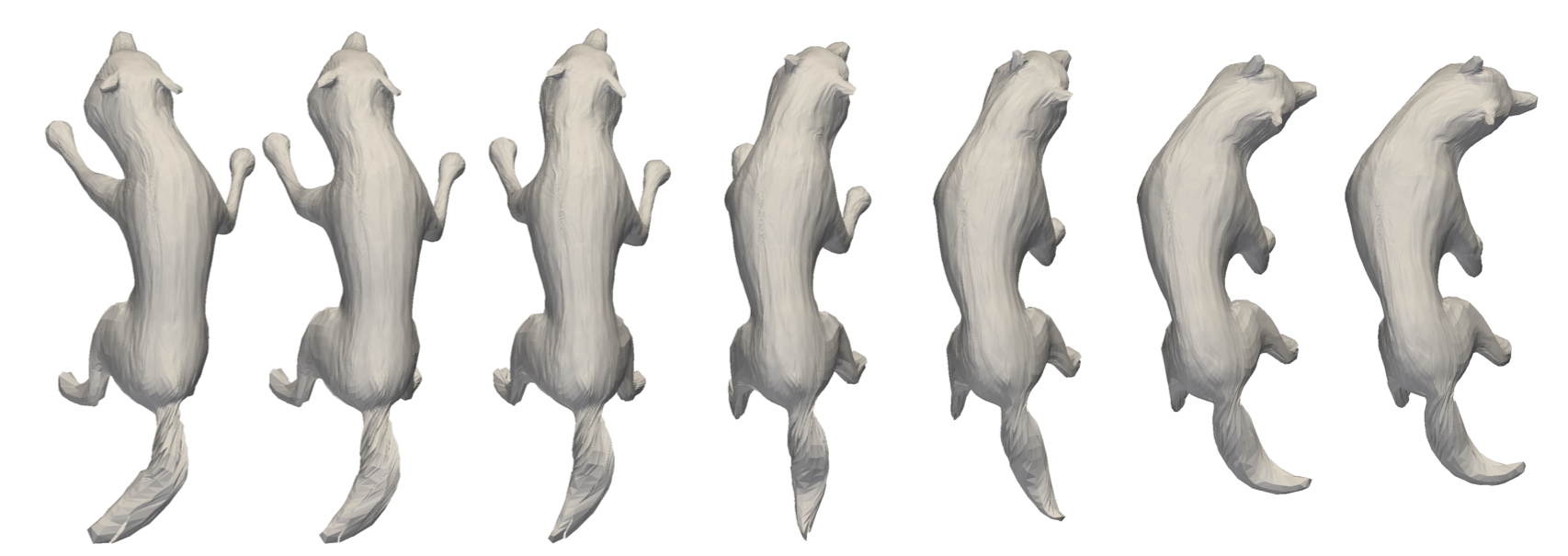}\\
    \textbf{(b)} Geodesic using our \MethodName~($2$ s).
    \caption{Geodesics comparison of numerical SRNF~\cite{laga2017numerical}  with the proposed \MethodName. Notice that the head and tail movement in \textbf{(b)} looks more natural compared to \textbf{(a)}.}
    \label{fig:animal_geodesics_2_suppl}
\end{figure}

\subsection{Summary statistics and shape generation}
\label{sec:summary_statistics}

Given a collection of 3D shapes $\{\surface_1,\dots, \surface_n\}$, the goal of statistical analysis is to compute the mean shape $\mean$ and the modes of variation, characterize the shape variability in the collection using a probability distribution, \eg a multivariate Gaussian, and finally sample from the distribution to generate novel 3D shapes. Since the pre-shape space $\surfaces$ is equipped with a nonlinear metric $\distance_{\surfaces}$, the mean shape can be computed by finding the shape  $\mean$ that is as close as possible, with respect to the metric 
$\distance_{\surfaces}$, to all the input samples:
\begin{equation}
    \mean = \argmin_{\surface \in \surfaces} \sum_{i=1}^{n} \distance_{\surfaces}\left(\surface, \surface_i\right)^2.
\label{equ:mean_shape}
\end{equation}

\noi This is computationally very expensive to solve as it involves shooting geodesics from $\surface$ to every sample $\surface_i$ a number of times (depending on the number of iterations required by the optimization algorithm).  We address this computational issue by first mapping the input surfaces to their SRNF representations $\{\srnf_1,\dots, \srnf_n\}$, performing Principal Component Analysis (PCA) in the SRNF space, and then mapping the computed statistical summaries back to the original space using the proposed \MethodName~framework. For instance, the mean $\mean_{\srnf}$ in the SRNF space is given by $\mean_{\srnf} =\frac{1}{n}\sum_{i=1}^{n}\srnf_i$.
Similarly, the modes of variation are the eigenvectors of the covariance matrix
$\cov = \frac{1}{n-1}\overset{n}{\underset{i=1}{\sum}} (\srnf_i - \mean_{\srnf}) (\srnf_i - \mean_{\srnf})^\top$. Let us denote by $\eigenvect_i$ and $\eigenval_i$ the $i-$th leading eigenvector and its corresponding eigenvalue. With this, the principal geodesic path $G_i$ that corresponds to the principal component $\eigenvect_i$ is given by:
\begin{equation}
    G_i(\tau) = \srnfmap^{-1}(\mean_{\srnf} + \tau \eigenvect_i), \quad \tau \in \real.
\end{equation}

\noi Here, $\srnfmap^{-1}$ refers to the SRNF inversion using the proposed \MethodName. Also,  with this formulation, any surface $\surface \in \surfaces$ can be written as:
\begin{equation}
    \surface = \srnfmap^{-1}\left( \mean_{\srnf} + \sum_{i=1}^k a_i \sqrt{\eigenval_i} \eigenvect_i\right),  \text{ with } a_i \in \real.
\label{eq:sampling}
\end{equation}

\noi Here, $k$ is the number of leading eigenvectors. Thus, one can generate a novel 3D surface by first sampling random values $\{a_i\}_{i=1}^k$ and using Eqn.~\eqref{eq:sampling} to generate a new surface $\surface$. To further control the quality of the generated surface, one can enforce the weights $\{a_i\}_{i=1}^k$ to be within a certain range, \eg $[-1, 1]$ to ensure that the generated surfaces are within one standard deviation around the mean.

%-------------------------------------------------------------------------

\begin{table}
    \centering
    \small
    \begin{tabular}{@{}lccc@{}}
    \toprule
    \textbf{FAUST~\cite{bogo2014faust}} & \textbf{Mean} $\downarrow$ & \textbf{Std} & \textbf{Median} \\
   % \midrule
   % \multicolumn{4}{c}{ FAUST~\cite{bogo2014faust}} \\
    \midrule
    LIMP~\cite{cosmo2020LIMP}  & 0.239 & 0.192 & 0.179 \\
    NeuroMorph~\cite{eisenberger2021neuromorph}  & 0.167 & 0.162 & 0.112 \\
    Ours  & \textbf{0.087} & \textbf{0.075} & \textbf{0.081} \\
    \bottomrule

    \textbf{DFAUST~\cite{dfaust:CVPR:2017}} & \textbf{Mean} $\downarrow$ & \textbf{Std} & \textbf{Median} \\
    %\multicolumn{4}{c}{ DFAUST~\cite{dfaust:CVPR:2017}}\\
    \midrule
    SAE~\cite{lemeunier2022SAE}  & 0.199 & 0.166 & 0.153 \\
    Ours  & \textbf{0.028} & \textbf{0.028} & \textbf{0.019} \\
    \bottomrule
    \end{tabular}
    \caption{We measure the pairwise geodesic error on five pairs from the FAUST~\cite{bogo2014faust}  and DFAUST~\cite{dfaust:CVPR:2017} dataset as the $\mathbb{L}^1$ distance between source and intermediate shape.}
    \label{tab:pairwise_geodesic}
\end{table}

\begin{table}
\centering

    \begin{adjustbox}{width=\columnwidth}

        \begin{tabular}{l|ccc}
        % \hline
        \textbf{} & FAUST & DFAUST & COMA \\
        \hline
        \textbf{LIMP (Geodesic error)}    & $3.82 \times 10^{-3}$ & $\mathbf{2.89 \times 10^{-4}}$ & $7.51 \times 10^{-4}$ \\
        \textbf{Ours (Geodesic error)}  & $\mathbf{2.96 \times 10^{-3}}$ & $1.14 \times 10^{-3}$ & $\mathbf{6.98 \times 10^{-4}}$ \\
        \hline
        \end{tabular}
        
    \end{adjustbox}
\caption{We measure the geodesic error on $100$ pairs from the FAUST~\cite{bogo2014faust}, DFAUST~\cite{dfaust:CVPR:2017}, and COMA~\cite{COMA:ECCV18} datasets using the LIMP~\cite{cosmo2020LIMP} evaluation metric.}
\label{tab:geodesic_errors}
\end{table}

\begin{figure}
    \centering
    \includegraphics[width=\linewidth,trim={0cm 0cm 0cm 0cm},clip ]{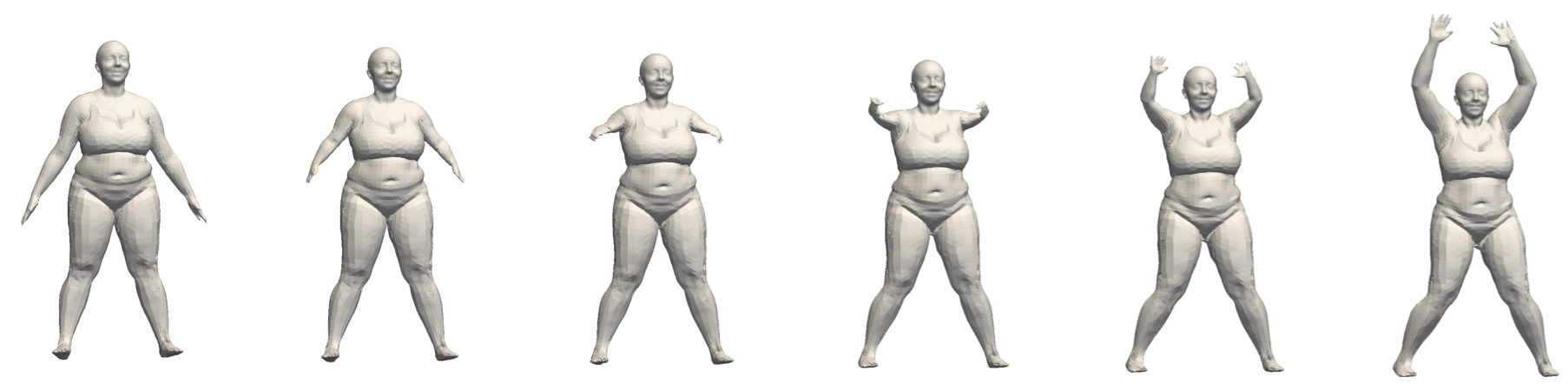}\\
    \textbf{(a)} Linear path.\\
    \includegraphics[width=\linewidth,trim={0cm 0cm 0cm 0cm},clip ]{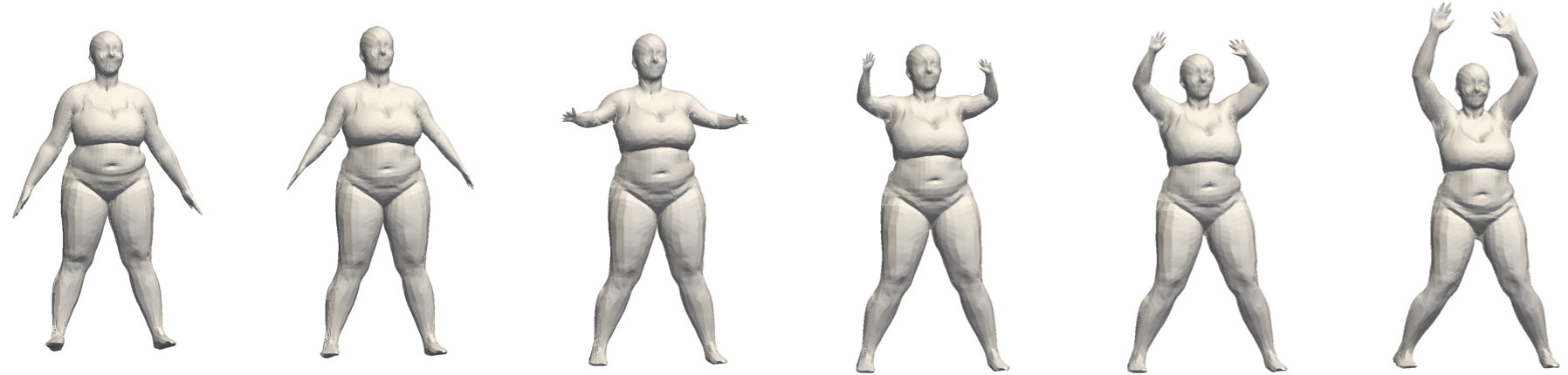}\\
    \textbf{(b)} Geodesic using numerical SRNF~\cite{laga2017numerical} ($818$ s).\\
    \includegraphics[width=\linewidth,trim={0cm 0cm 0cm 0cm},clip ]{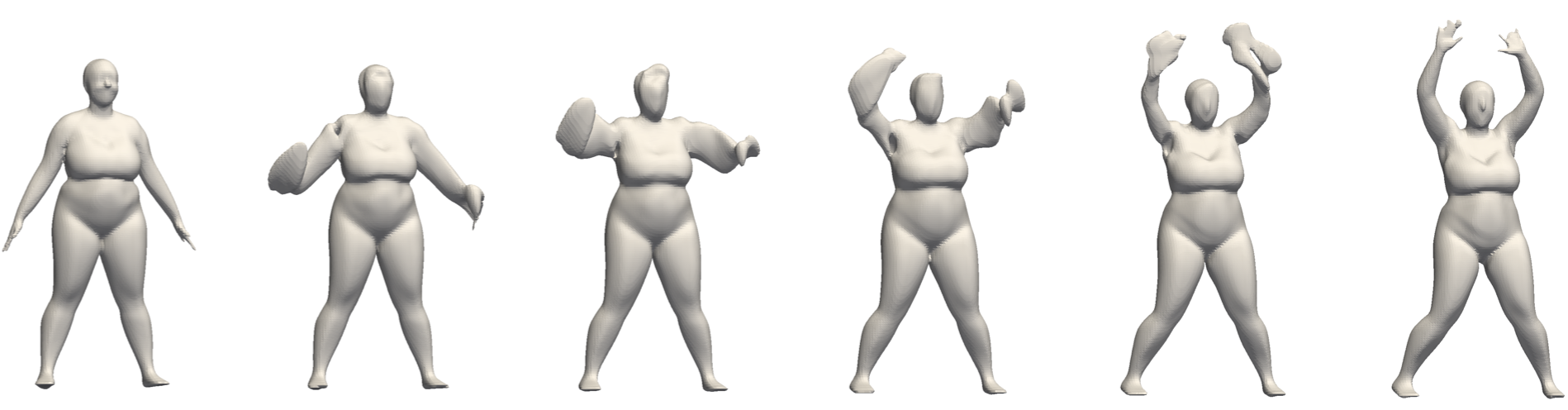}\\
    \textbf{(c)} Geodesic using implicit neural surface~\cite{sang2025implicit} (authors reported a time of $20$ min; time taken on our machine is $2$ h).\\
    \includegraphics[width=\linewidth,trim={0cm 0cm 0cm 0cm},clip ]{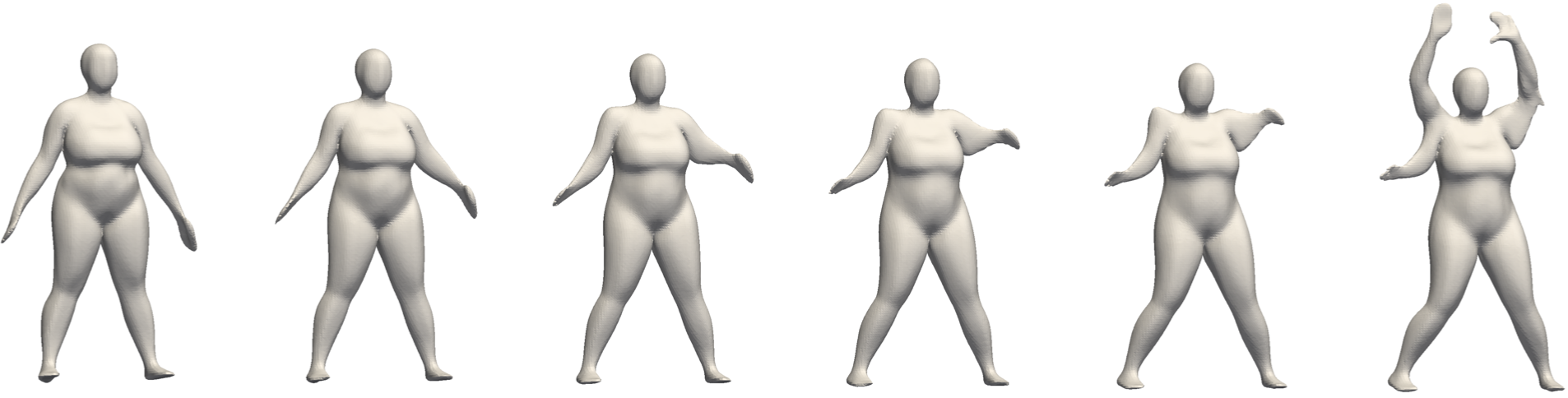}\\
    \textbf{(d)} Geodesic using 4Deform~\cite{sang20254deform} (authors reported a time of $7$ min; time taken on our machine is $1$ h).\\
    \includegraphics[width=\linewidth,trim={0cm 0cm 0cm 0cm},clip ]{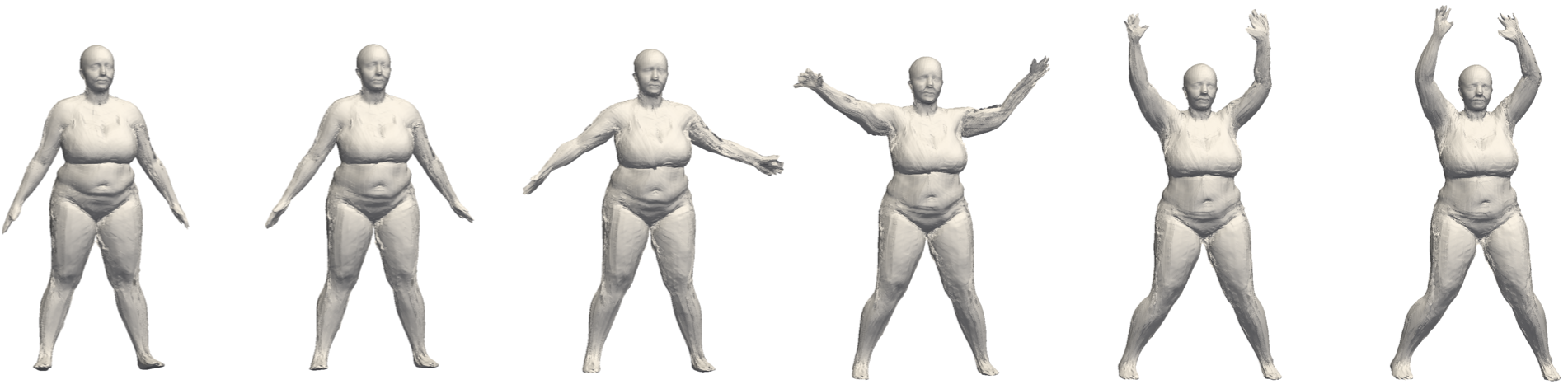}\\
    \textbf{(e)} Geodesic  using our \MethodName\ inversion ($3$ s).\\
    \caption{Comparison of the geodesic computed using the proposed \MethodName\ with state-of-the-art methods.}
    \label{fig:human_geodesics_implicit_comparison_1}
\end{figure}

\section{Results}
\label{sec:results}

\noi\textbf{Datasets.} We have tested the \MethodName\ framework on a wide range of 3D and 4D datasets covering faces, humans, and animal surfaces from MPI COMA~\cite{COMA:ECCV18}, MPI FAUST~\cite{bogo2014faust}, MPI DFAUST~\cite{dfaust:CVPR:2017}, MPI CAPE~\cite{CAPE:CVPR:20}, and Animal3D~\cite{xu2023animal3d} datasets. We  spherically parameterize all the 3D surfaces using  Kurtek \etal's implementation~\cite{kurtek2013landmark} of the approach of Hoppe and Praun~\cite{praun2003spherical}.

% All the datasets come with registered triangulation. 

% \begin{figure}
%     \centering
%     \includegraphics[width=\linewidth,trim={0cm 0cm 0cm 0cm},clip ]{figures/spatiotemporal/coma/COMA_bareteethsource.png}\\
%     \textbf{(a)} Source.\\
%     \includegraphics[width=\linewidth,trim={0cm 0cm 0cm 0cm},clip ]{figures/spatiotemporal/coma/COMA_bareteethtarget_unregistered.png}\\
%     \textbf{(b)} Target before spatiotemporal registration.\\
    
%     \includegraphics[width=\linewidth,trim={0cm 0cm 0cm 0cm},clip ]{figures/spatiotemporal/coma/COMA_bareteethtarget_registered.png}\\
%    \textbf{(c)} Target after spatiotemporal registration.
%     \caption{Spatiotemporal registration of two 4D faces from the COMA dataset.}
%     % with a $50$ times faster speedup; see the  Supplementary Material for additional results.}
%     \label{fig:coma_1_spatiotemporal}
% \end{figure}

\begin{figure}
    \centering
    \includegraphics[width=\linewidth]{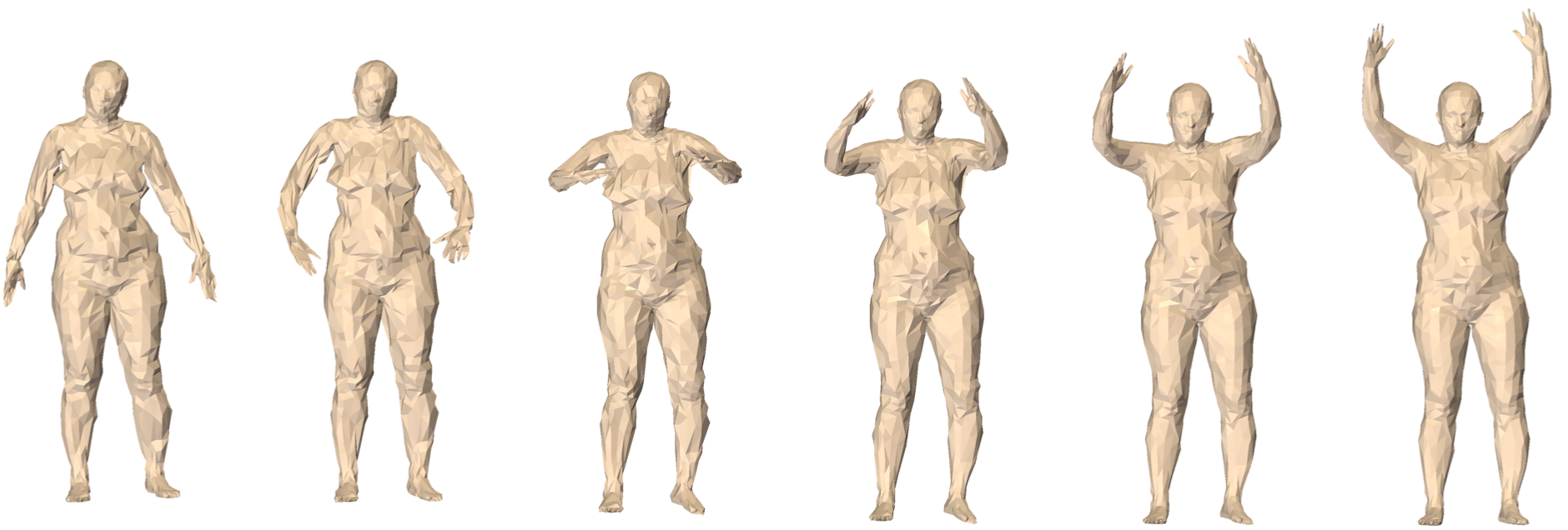}\\
    \textbf{(a)} Geodesic using LIMP~\cite{cosmo2020LIMP} ($1$ s).\\
    \includegraphics[width=\linewidth]{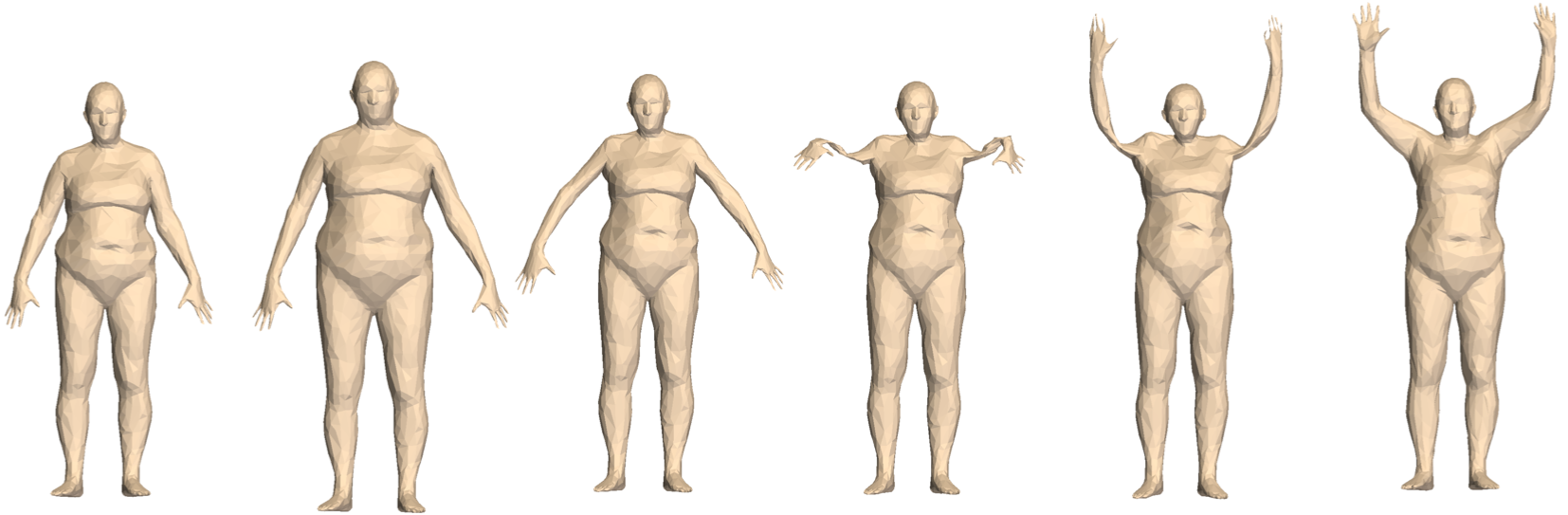}\\
    \textbf{(b)} Geodesic using NeuroMorph~\cite{eisenberger2021neuromorph} ($1$ s).\\
    \includegraphics[width=\linewidth]{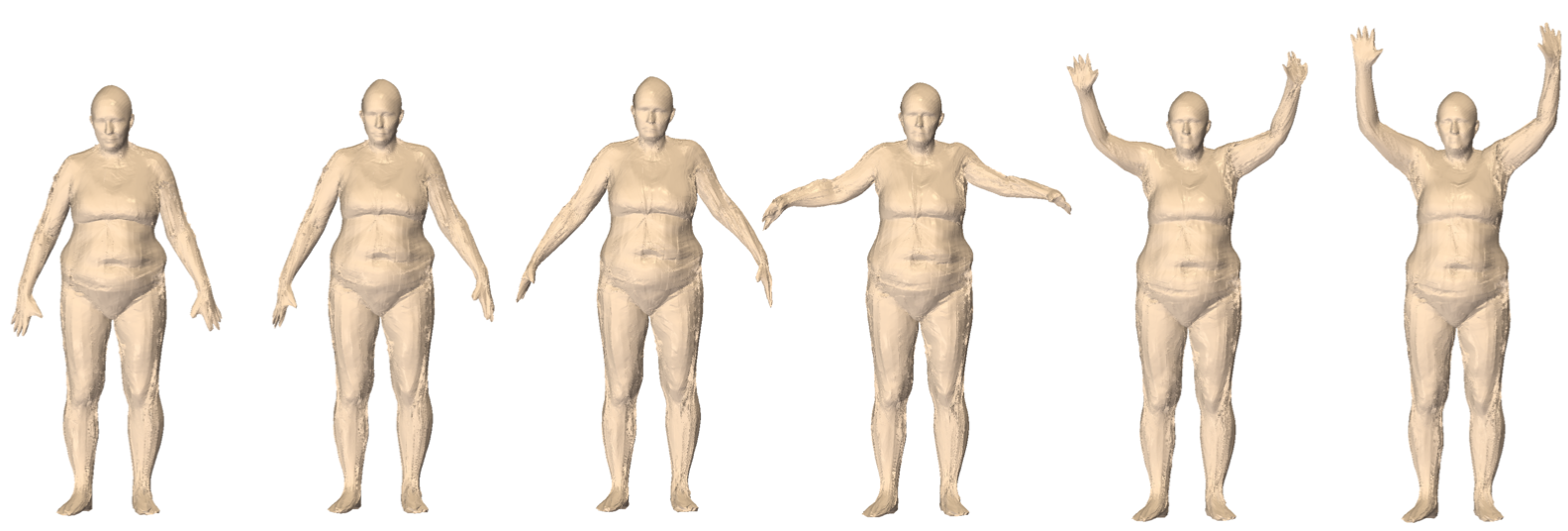}\\
    \textbf{(c)} Geodesic using our \MethodName\ inversion ($3$ s).\\
    
    \caption{Examples of geodesics generated using the discrete methods \textbf{(a)} LIMP~\cite{cosmo2020LIMP} and \textbf{(b)} NeuroMorph~\cite{eisenberger2021neuromorph} and compared with \textbf{(c)} our continuous \MethodName. }
    \label{fig:limp_1_geodesics_main}
\end{figure}

\noi\textbf{Implementation details.} The \MethodName\ framework was trained on an NVIDIA GeForce RTX 4090 GPU with an Intel Core i9 2.4 GHz processor. The proposed SRNF inversion network was optimized using AdamW~\cite{Ilya2017ADAMW} with a CosineAnnealing~\cite{Ilya2017CosineAnnealing} learning rate scheduler starting from $1e^{-4}$ and gradually reduced to $1e^{-6}$. We train the network for $1,000$ epochs with a batch size of $16$, which takes around $8$ to $24$ hours to train, depending on the complexity of the dataset. For each batch, we sample a $16 \times 16$ patch of 3D points to update the network parameters. We follow this for the human, animal, and face datasets.

\begin{figure*}[t]
    \includegraphics[width=\linewidth,trim={1cm 10.2cm 0cm 3cm},clip ]{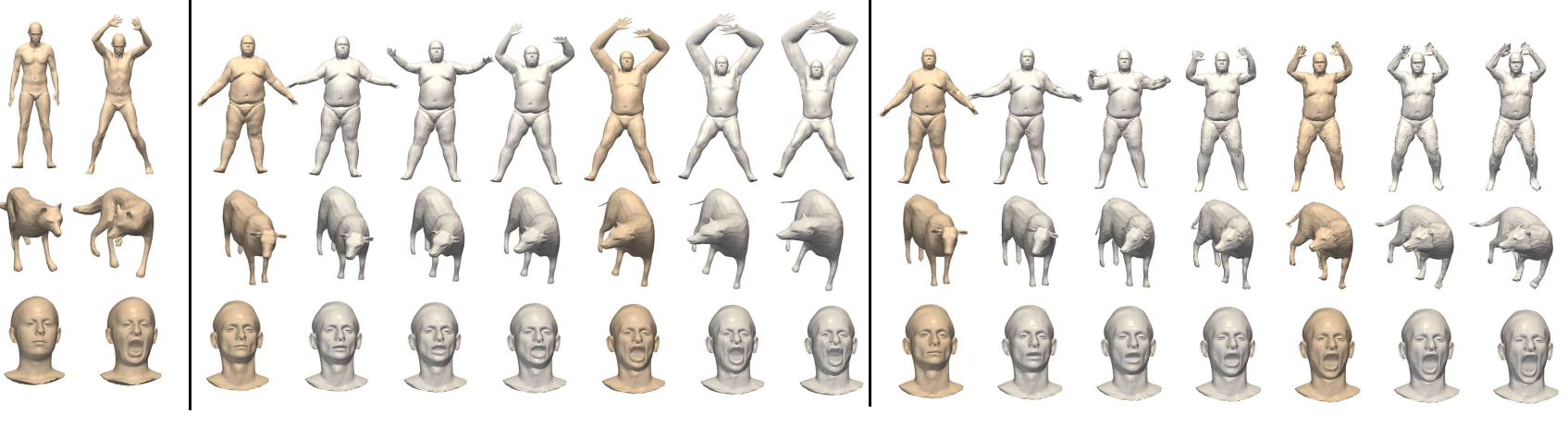}\\
    \begin{adjustbox}{width=\textwidth}
    \begin{tabular}{@{\hspace{2.8cm}} l l l l l l l r r r r r r r }
    $\alpha(0)$&$\alpha(0.25)$&$\alpha(0.5)$&$\alpha(0.75)$&$\alpha(1)$&$\alpha(1.25)$&$\alpha(1.5)$&$\alpha(0)$&$\alpha(0.25)$&$\alpha(0.5)$&$\alpha(0.75)$&$\alpha(1)$&$\alpha(1.25)$&$\alpha(1.5)$
    \end{tabular}
    \end{adjustbox}

    \begin{tabular}{@{\hspace{0.3cm}} l @{\hspace{0.65cm}} l @{\hspace{4.4cm}} c @{\hspace{3cm}} c @{\hspace{4.3cm}} c}
        $\surfaceone \to h_1$ & $\surfacetwo$ & $h_2$ & $\surfacetwo$ & $h_2$ \\
        (a) Source.  & \multicolumn{2}{c}{ (b) Using numerical SRNF~\cite{laga2017numerical}.} & \multicolumn{2}{c}{ \quad (c) Using the proposed method.}
    \end{tabular}
    
    \caption{Examples of deformation transfer for humans, animals, and faces.  \textbf{(a)} The source deformation, \textbf{(b)} deformation transfer by linear interpolation and extrapolation in the space of SRNF $\srnfs$ followed by SRNF inversion of $\surfaces$ using numerical SRNF~\cite{laga2017numerical}, and \textbf{(c)} using the proposed \MethodName.}
    \label{fig:deformation_transfer_three_examples_with_extrapolation}
\end{figure*}

\begin{figure}
    \centering
    \includegraphics[width=0.95\linewidth,trim={0.4cm 2cm 0cm 2cm},clip ]{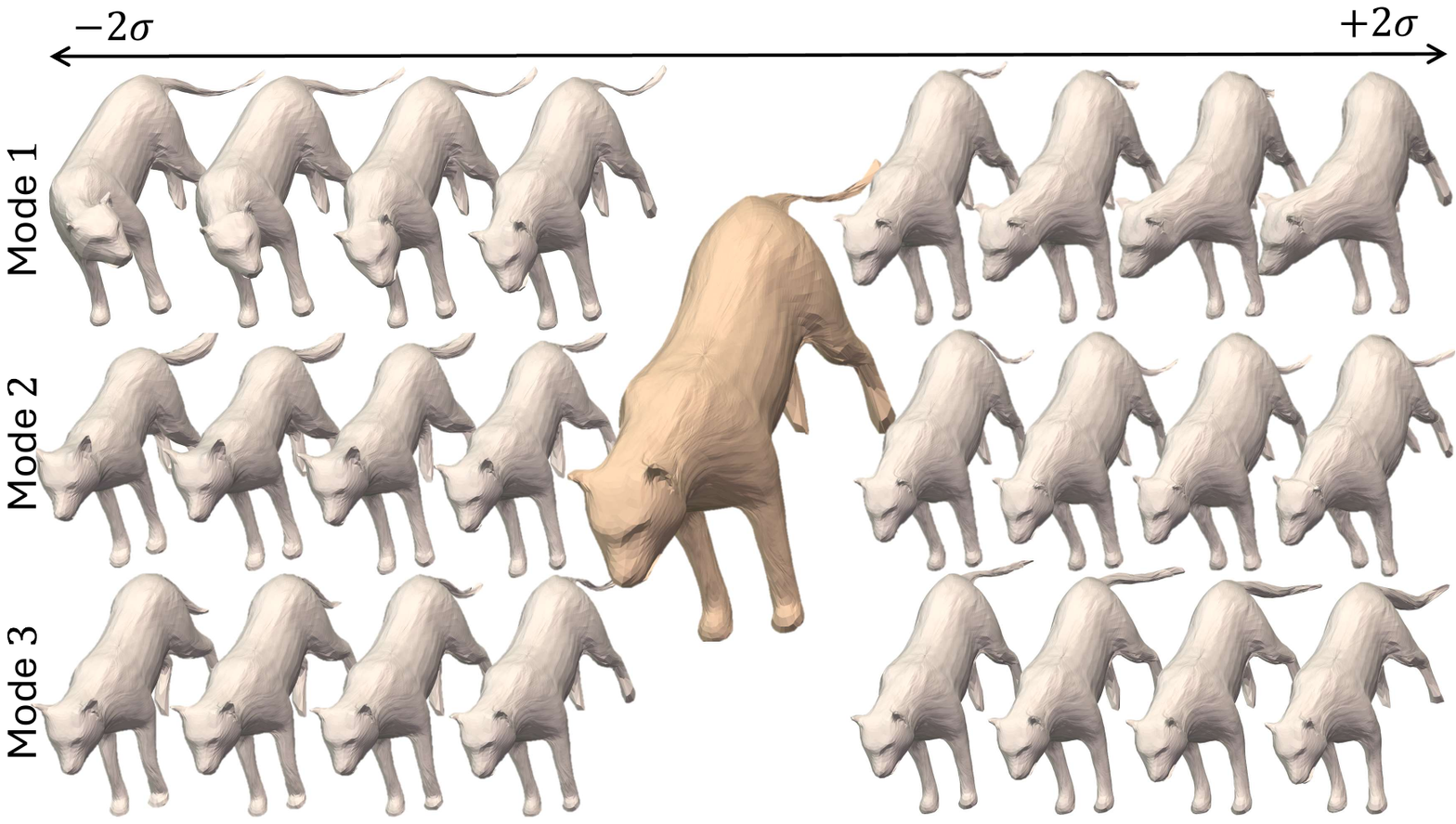}\\
    (a) Summary statistics of animals.\\
    \includegraphics[width=0.98\linewidth,trim={0.4cm 2cm 0cm 2cm},clip ]{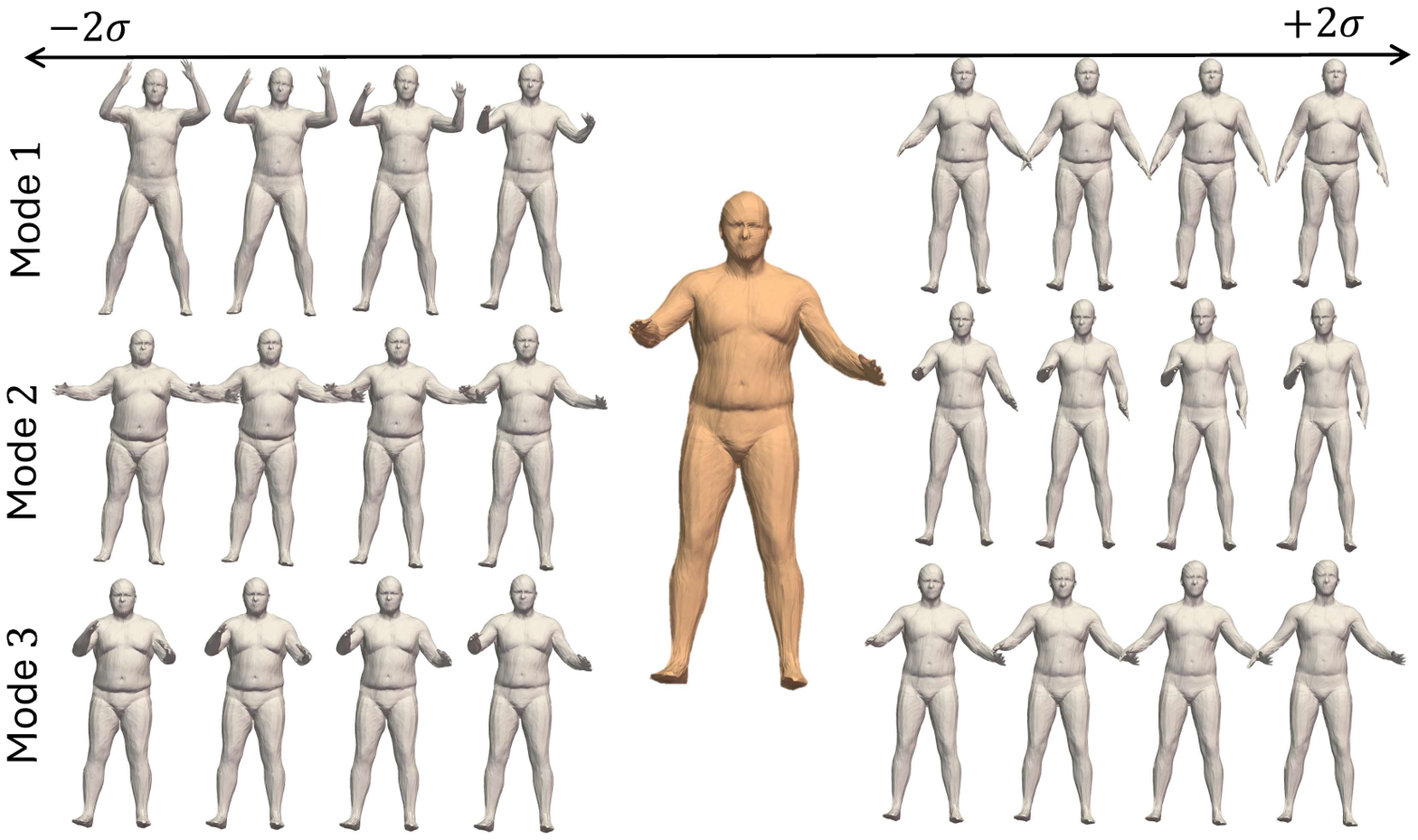}
    \\
    (b) Summary statistics of humans.
    \caption{Summary statistics computed with the proposed \MethodName~on the animal and human datasets.}

    \label{fig:animal_human_summary_statistics}
\end{figure}

\begin{figure}
    \centering
    \includegraphics[width=\linewidth,trim={0.9cm 0cm 0cm 0cm},clip ]{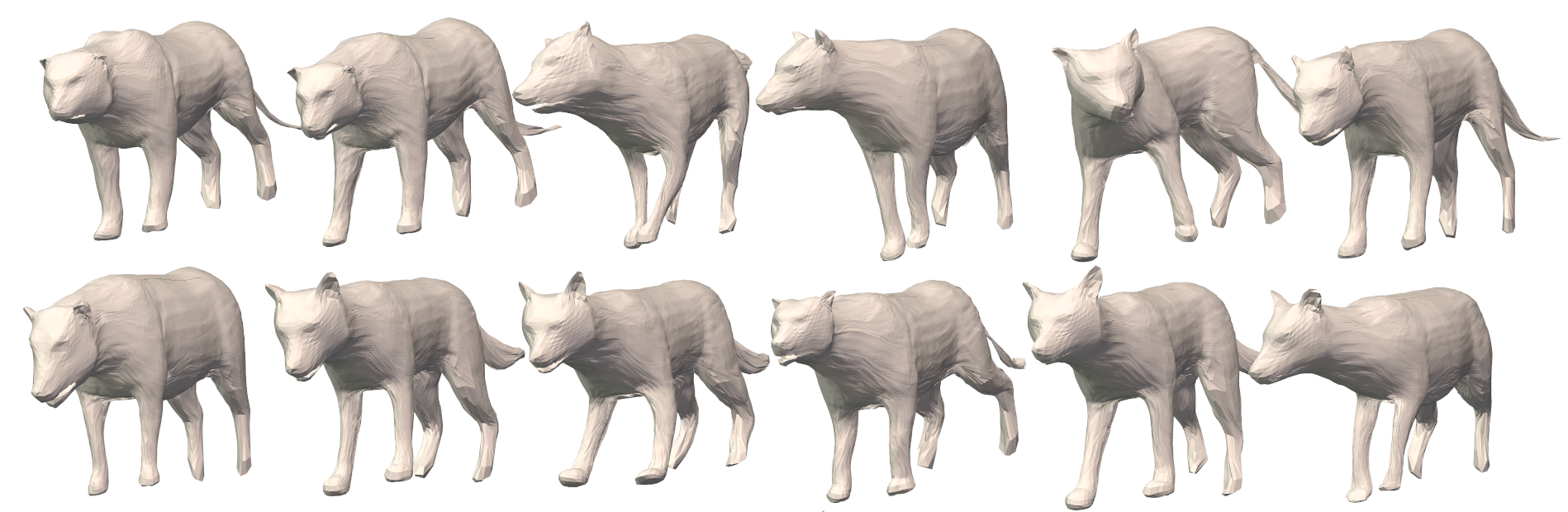}
    
    \caption{$12$ randomly generated 3D animals with \MethodName.}
    \label{fig:animal_shape_generation}
\end{figure}

\begin{figure}
    \centering
    % \caption{Summary statistics computed using the proposed \MethodName~on the human dataset.}

    \includegraphics[width=\linewidth,trim={0.5cm 5cm 0cm 9.5cm},clip ]{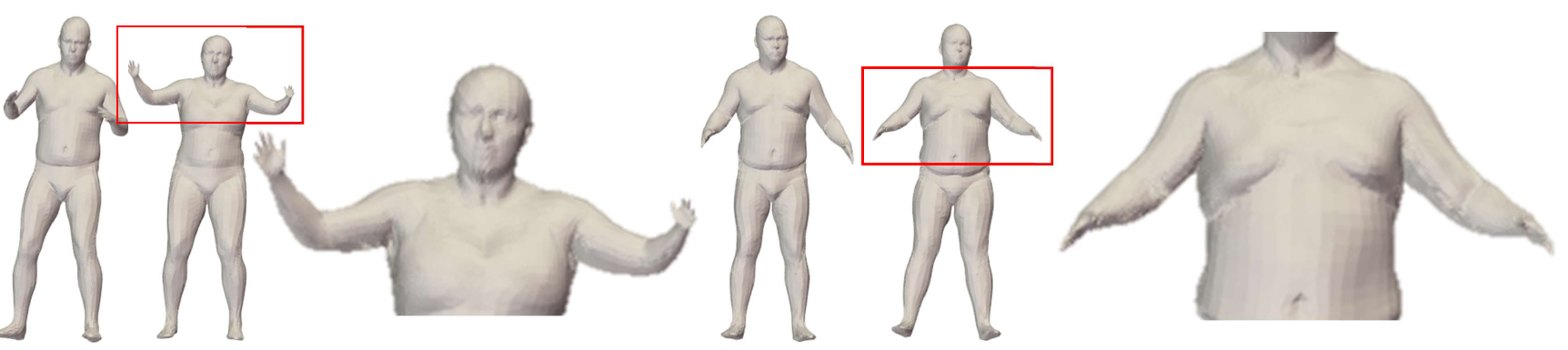}
    \\
    (a) Randomly generated humans using numerical SRNF~\cite{laga2017numerical}.
    \includegraphics[width=\linewidth,trim={0.5cm 11.5cm 0cm 2.75cm},clip ]{figures/statistics/Human/random_samples_human_comparison_2.pdf}
    \\
    (b) Randomly generated humans using our \MethodName.
    \caption{Generated 3D shapes with two close-up views using \textbf{(a)} numerical SRNF~\cite{laga2017numerical} and \textbf{(b)} our \MethodName.}
    % \caption{\textbf{(a)} Summary statiscis computed using the proposed \MethodName~on the human dataset. \textbf{(b)} and \textbf{(c)} shows 3D shape generation with two close-up views using  numerical SRNF~\cite{laga2017numerical} and \textbf{(c)} our \MethodName.}

    \label{fig:human_shape_generation_comparison}
\end{figure}

\subsection{SRNF inversion}

% \begin{wrapfigure}{r}{0.6\textwidth}
%     \centering
%     \includegraphics[width=\linewidth,trim={0cm 5.25cm 5cm 2.5cm},clip ]{figures/srnf_quality/neural_srnf_main_figure_3_pyvista.pdf}
%     \begin{adjustbox}{width=\textwidth}
%         \begin{tabular}{@{\hspace{0.5cm}} l @{\hspace{0.25cm}} l @{\hspace{0.25cm}} c @{\hspace{0.25cm}} c @{\hspace{2cm}} l @{\hspace{1cm}} c @{\hspace{0.25cm}} c @{\hspace{10cm}}}
%          (a) GT.  & (b) NumSRNF~\cite{laga2017numerical}. &  (c) SAE~\cite{lemeunier2022SAE} & (d) Ours. &  (a) GT.  & (b) NumSRNF~\cite{laga2017numerical}. & (d) Ours\\
%         \end{tabular}    
%     \end{adjustbox}
    
%     \caption{Accuracy of the SRNF inversion on 3D humans and animals. The pointwise error is shown as a heatmap for (b), (c), and (d).}
    
%     % \textbf{(a)} The ground-truth surfaces. \textbf{(b)} Reconstructed surfaces using the numerical SRNF~\cite{laga2017numerical}.  \textbf{(c)}  Reconstructed surfaces using  our \MethodName~method. The pointwise error is shown as a heatmap for (b) and (c). The Supplementary Material provides additional results.}
%     \label{fig:neural_srnf_quality_mainl}
% \end{wrapfigure}

We first evaluate, quantitatively and qualitatively, the accuracy and performance of the SRNF inversion procedure proposed in this paper and compare it against the numerical inversion procedure proposed by Laga \etal~\cite{laga2017numerical}. 
For this experiment, we use test surfaces from the COMA, CAPE, DFAUST, Animal with tail, and Animal without tail datasets. For each dataset, we randomly sample $100$ test surfaces and convert each surface to the SRNF representation. We then apply the inversion frameworks and report the average of the mean squared distance $\ltwo$ between the ground-truth and the recovered surfaces using our method and the numerical SRNF method~\cite{laga2017numerical}.

Table~\ref{tab:comparison} reports the mean, median, and standard deviation of the inversion error and inference time for the five datasets. As one can see, our inversion framework achieves superior performance on all datasets across the three metrics and is able to perform SRNF inversion within $0.10$ to  $0.65$ s per surface compared to $4.73$ to $69$ s for~\cite{laga2017numerical}. Thus, our proposed framework is $50$ to $100$ times faster than the traditional numerical SRNF~\cite{laga2017numerical} while achieving significantly better accuracy. % Note that the errors are reported in ($\times 10^{-3}$) and the inference times are for a single 3D surface.

% \begin{figure}%[tb]
%     \centering
%     \includegraphics[width=\linewidth,trim={0cm 0cm 0cm 0cm},clip ]{figures/geodesics/implicit_comparison/implicit_neural_surface_result.png}\\
%     \small{(a) Using~\cite{sang2025implicit} (\cite{sang2025implicit} reported a time of $20$ min, time taken on our machine is $2$ hrs)}.\\
%     \includegraphics[width=\linewidth,trim={0cm 0cm 0cm 0cm},clip ]{figures/geodesics/implicit_comparison/4deform_results_implicit_neural_surface.png}\\
%     \small{(b) Using 4Deform~\cite{sang20254deform} (\cite{sang20254deform} reported $7$ min, time taken on our machine is $1$ hr)}.\\
%     \includegraphics[width=\linewidth,trim={0cm 0cm 0cm 0cm},clip ]{figures/geodesics/implicit_comparison/ours_implicit_neural_surface_results.png}\\
%     \small{\textbf{(c)} Using  our \MethodName\ ($3$ s)}.
%     \caption{Geodesics generated with our method and the state of the art. } 
%     % The Supplementary Material provides additional comparisons.} 
%     \label{fig:human_geodesics_implicit_comparison_1}
% \end{figure}

% Figure~\ref{fig:neural_srnf_quality_mainl} demonstrates the qualitative results of the SRNF inversion on the humans and animals datasets. In this figure, we display \textbf{(a)} the ground-truth 3D surface, \textbf{(b)} the inversion computed using the numerical inversion~\cite{laga2017numerical}, \textbf{(c)} the inversion of spectral basis computed using the autoencoder~\cite{lemeunier2022SAE}, and \textbf{(d)} the inversion computed using \MethodName. 
Figure~\ref{fig:neural_srnf_quality_mainl} demonstrates the qualitative results of the inversion on 3D humans and animals. In this figure, we display \textbf{(a)} the ground-truth 3D surface, \textbf{(b)} the inversion of SRNF computed using the proposed \MethodName, \textbf{(c)} the inversion of SRNF computed using the numerical inversion~\cite{laga2017numerical}, and \textbf{(d)} the inversion of spectral basis computed using the SAE~\cite{lemeunier2022SAE}. We also plot the reconstruction error, \ie the pointwise error between the original surface and the reconstructed surface obtained by inversion. As shown in Figure~\ref{fig:neural_srnf_quality_mainl}-\textbf{(b)}, the inversion error of our method is signifcantly lower than that of the state of the art across all regions of the surface, including at the extremities (such as fingers) and non-convex regions.

%with an error that is less than $0.1$. Note that for each surface,  \MethodName's  errors are between $0$ and $0.1$ (red represents a higher error and gray represents no error).

Figure~\ref{fig:neural_srnf_quality_suppl} shows additional qualitative results of the proposed \MethodName~inversion on 3D models of human bodies, faces, and animals. We show the error as a heatmap computed using the approach discussed in Figure~\ref{fig:neural_srnf_quality_mainl}.

\subsection{Geodesics}
\label{sec:results_geodesics}

% Ideally, this should be zero. The mean, median, and standard deviation of the error for LIMP  are 0.239, 0.179, and 0.192, respectively, whereas our method attains lower errors of 0.0876, 0.0818, and 0.0756.

% report the quantitative comparison by assessing whether the geodesic distances  between points along the surface of the 3D models are preserved when interpolating  3D models that differ in their pose. We

% \begin{figure}
%     \centering
%     \includegraphics[width=\linewidth,trim={0cm 0cm 0cm 0cm},clip ]{figures/spatiotemporal/dfaust/DFAUST_jumping_source.png}\\
%     \textbf{(a)} Source.\\
%     \includegraphics[width=\linewidth,trim={0cm 0cm 0cm 0cm},clip ]{figures/spatiotemporal/dfaust/DFAUST_Jumping_videotarget_unregistered.png}\\
%     \textbf{(b)} Target before temporal registration.\\
    
%     \includegraphics[width=\linewidth,trim={0cm 0cm 0cm 0cm},clip ]{figures/spatiotemporal/dfaust/DFAUST_Jumping_videotarget_registered.png}\\
%    \textbf{(c)} Target after temporal registration.
%     \caption{Temporal registration of two 4D humans from the DFAUST dataset performing a jumping action at different speeds.}
%     \label{fig:human_1_spatiotemporal}
% \end{figure}

\begin{figure}

    \centering
    \includegraphics[width=\linewidth,trim={0cm 0cm 0cm 0cm},clip ]{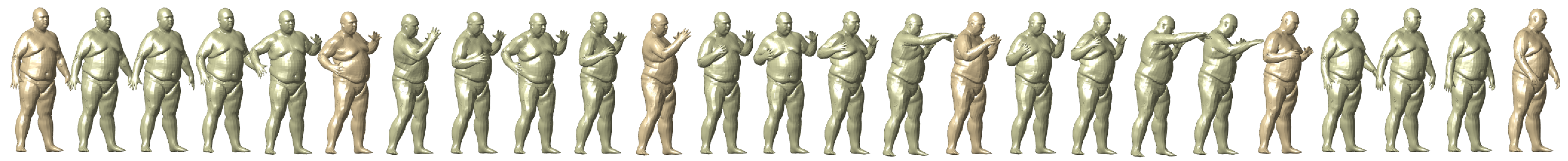}\\
    \textbf{(a)} Source.\\
    \includegraphics[width=\linewidth,trim={0cm 0cm 0cm 0cm},clip ]{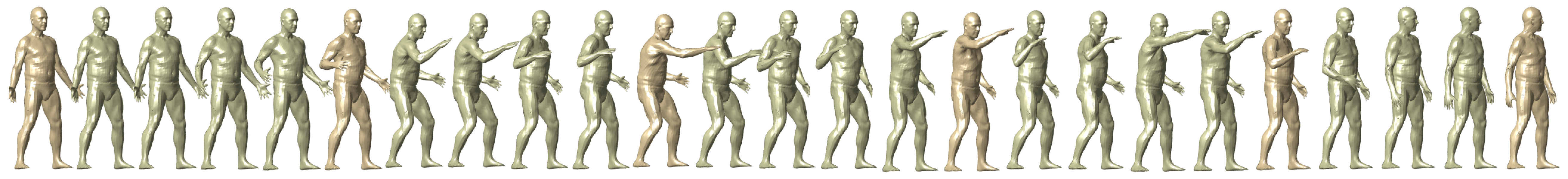}\\
    \textbf{(b)} Target before temporal registration.\\
    \includegraphics[width=\linewidth,trim={0cm 0cm 0cm 0cm},clip ]{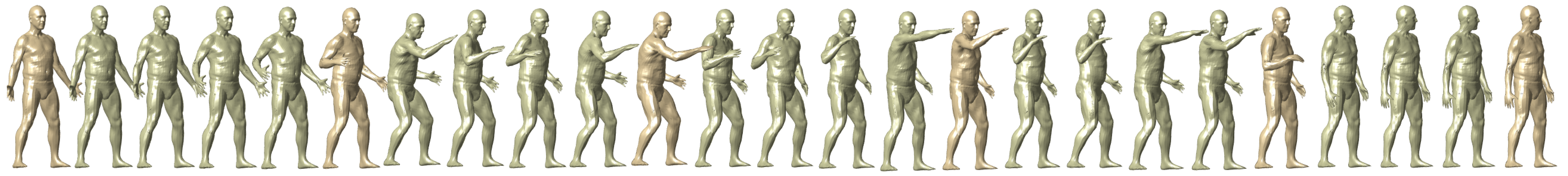}\\
   \textbf{(c)} Target after temporal registration.
   
    \rule{\linewidth}{0.5pt}\\

    % \includegraphics[width=0.98\linewidth,trim={0cm 0cm 0cm 0cm},clip ]{figures/DFAUST_geodesics_temporal/DFAUST_jumping_video_linear_geod_reg_1.png}\\
    % \includegraphics[width=0.98\linewidth,trim={0cm 0cm 0cm 0cm},clip ]{figures/DFAUST_geodesics_temporal/DFAUST_jumping_video_srnf_geod_unreg_2.png}\\

    %  {\color{black}
    % \fbox{%
    %     \includegraphics[width=0.98\linewidth,trim={0cm 0cm 0cm 0cm},clip ]    {figures/DFAUST_geodesics_temporal/DFAUST_jumping_video_srnf_geod_unreg_3.png}%
    % }}\\
    % \includegraphics[width=0.98\linewidth,trim={0cm 0cm 0cm 0cm},clip ]{figures/DFAUST_geodesics_temporal/DFAUST_jumping_video_srnf_geod_unreg_4.png}\\
    % \includegraphics[width=0.98\linewidth,trim={0cm 0cm 0cm 0cm},clip ]{figures/DFAUST_geodesics_temporal/DFAUST_jumping_video_linear_geod_unreg_5.png}\\

    % 4D geodesic using \MethodName~before temporal registration.

    % \rule{\linewidth}{0.5pt}\\
    
    \centering
    \includegraphics[width=0.98\linewidth,trim={0cm 0cm 0cm 0cm},clip ]{figures/DFAUST_geodesics_temporal/DFAUST_jumping_video_linear_geod_reg_1.png}\\
    \includegraphics[width=0.98\linewidth,trim={0cm 0cm 0cm 0cm},clip ]{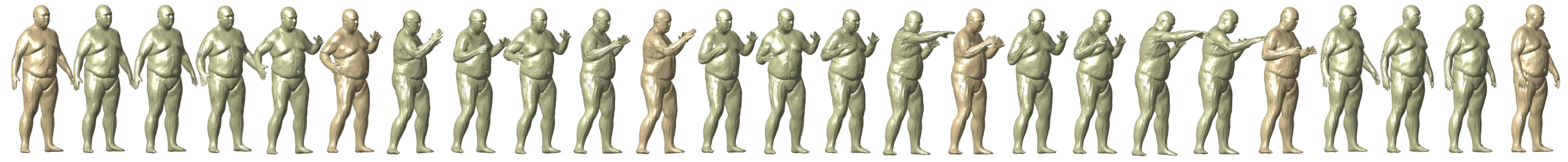}\\

     {\color{black}
    \fbox{%
        \includegraphics[width=0.98\linewidth,trim={0cm 0cm 0cm 0cm},clip ]    {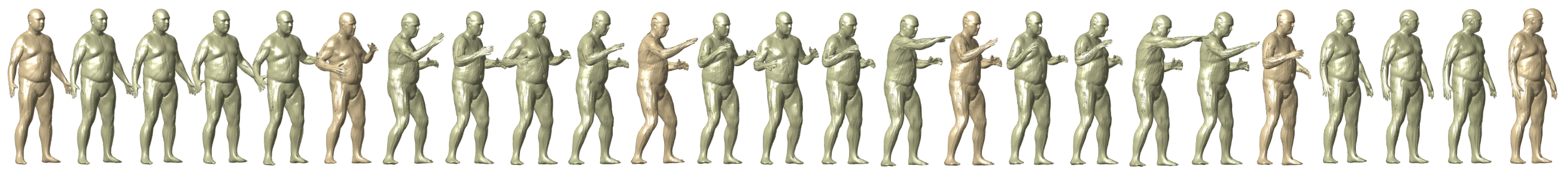}%
    }}\\
    \includegraphics[width=0.98\linewidth,trim={0cm 0cm 0cm 0cm},clip ]{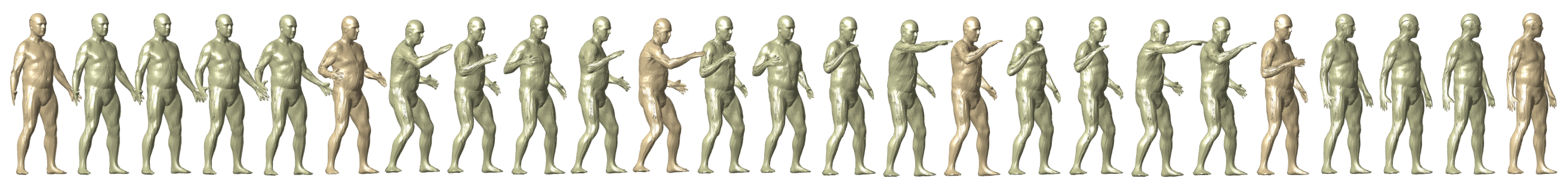}\\
    \includegraphics[width=0.98\linewidth,trim={0cm 0cm 0cm 0cm},clip ]{figures/DFAUST_geodesics_temporal/DFAUST_jumping_video_linear_geod_reg_5.png}\\

    % 4D geodesic using \MethodName~after temporal registration.

    % \rule{\linewidth}{0.5pt}\\

    \caption{Example of the temporal registration and geodesic generated after temporal alignment of source and target 4D surfaces using the proposed \MethodName~framework. The mean 4D surface is highlighted with a black rectangle. Note that the geodesics look more natural and are generated within $15$ s, which is $180$ times faster than the numerical SRNF~\cite{laga2017numerical}; see the  Supplementary Material for additional results. }
    % with a $50$ times faster speedup; see the  Supplementary Material for additional results.}
    \label{fig:spatiotemporal_geodesics_neural_srnf}
\end{figure}

\begin{figure*}
    \centering
    \includegraphics[width=0.91\linewidth,trim={0cm 0cm 0cm 0cm},clip ]{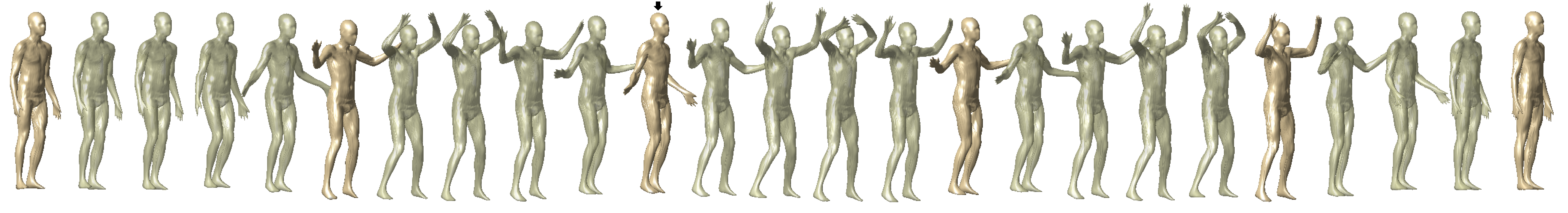}\\
    \textbf{(a)} Source.\\
    \includegraphics[width=0.91\linewidth,trim={0cm 0cm 0cm 0cm},clip ]{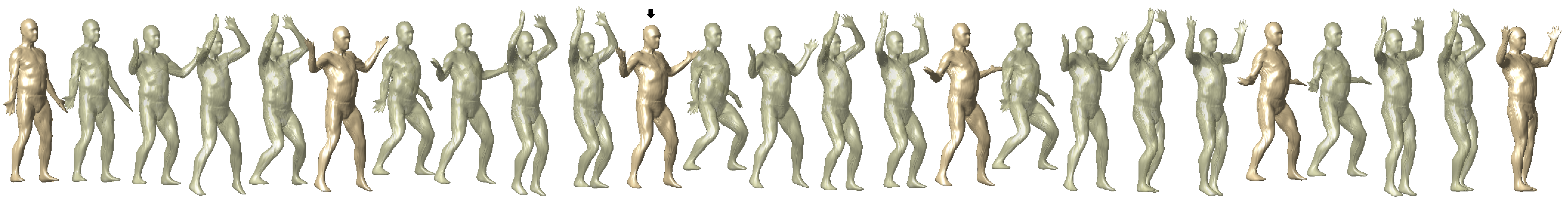}\\
    \textbf{(b)} Target before temporal registration.\\
    \includegraphics[width=0.91\linewidth,trim={0cm 0cm 0cm 0cm},clip ]{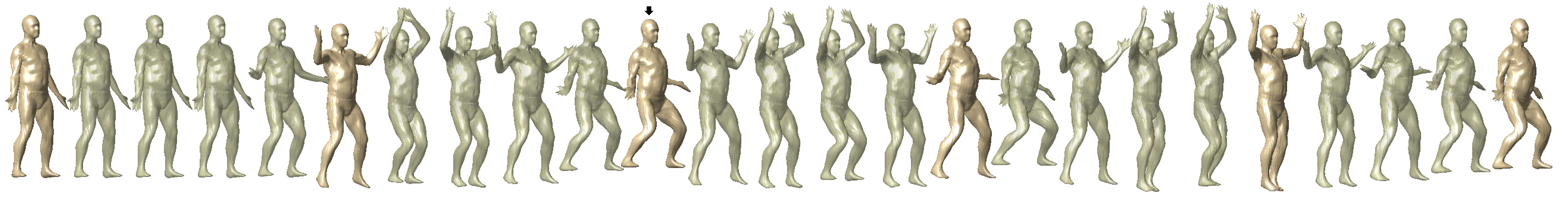}\\
    \textbf{(c)} Target after temporal registration.

    \includegraphics[width=0.91\linewidth,trim={0cm 0cm 0cm 0cm},clip ]{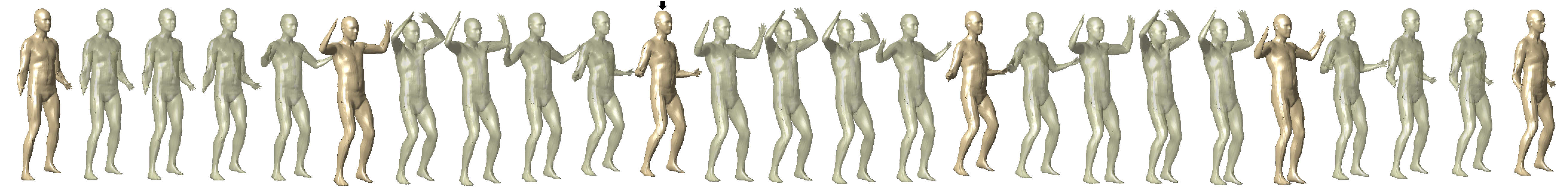}\\
    \textbf{(d)} Mean (with numerical SRNF~\cite{laga2017numerical}, after temporal registration, time taken $45$ min).\\

    \includegraphics[width=0.91\linewidth,trim={0cm 0cm 0cm 0cm},clip ]{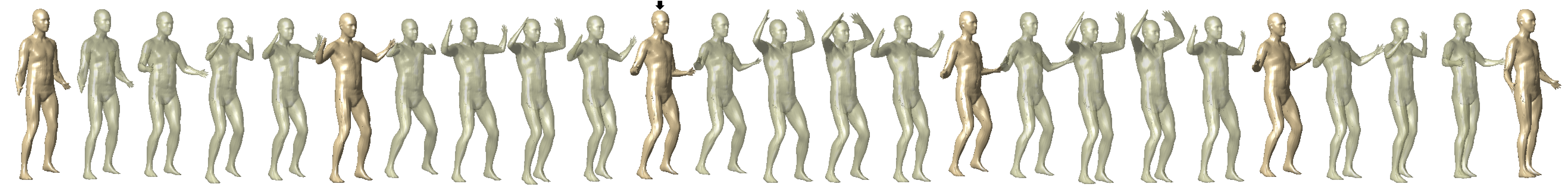}\\
    \textbf{(e)} Mean (with numerical SRNF~\cite{laga2017numerical}, before temporal registration, time taken $45$ min).

    \includegraphics[width=0.91\linewidth,trim={0cm 0cm 0cm 0cm},clip ]{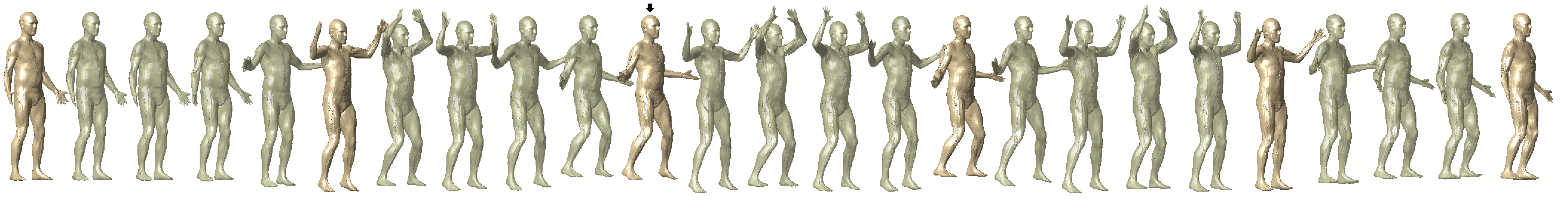}\\
    \textbf{(f)} Mean (with \MethodName, after temporal registration, time taken $15$ s).

    \includegraphics[width=0.91\linewidth,trim={0cm 0cm 0cm 0cm},clip ]{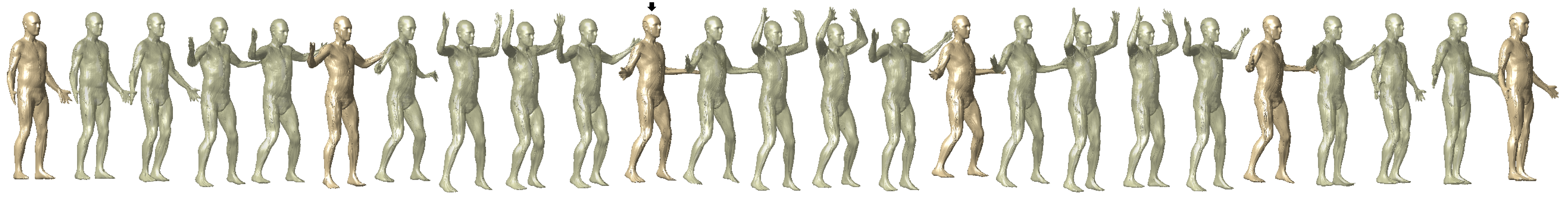}\\
    \textbf{(g)} Mean (with \MethodName, before temporal registration, time taken $15$ s).

    \includegraphics[width=0.75\linewidth,trim={0cm 0cm 0cm 0cm},clip ]{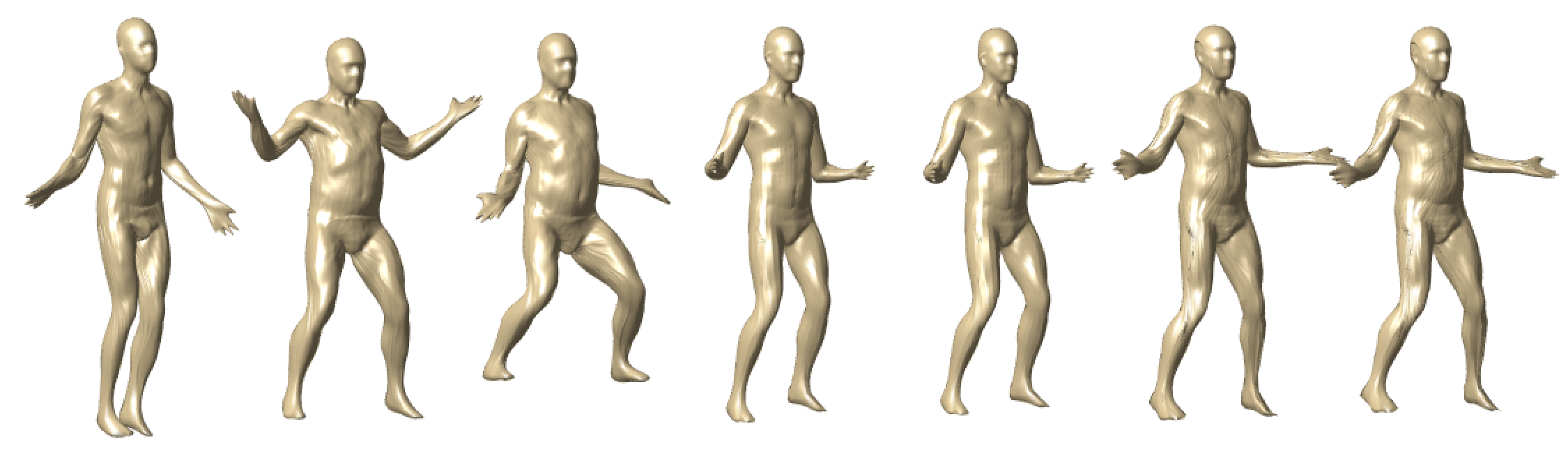}\\

    \noindent
\hspace*{2pt}%
\begin{adjustbox}{width=\dimexpr\textwidth-2pt\relax}
    \scriptsize
    \begin{tabular}{
        @{\hspace{3.4cm}}
        c @{\hspace{1.2cm}}
        c @{\hspace{1.0cm}}
        c @{\hspace{0.8cm}}
        c @{\hspace{0.7cm}}
        c @{\hspace{0.8cm}}
        c @{\hspace{1cm}}
        c
        @{\hspace{3.2cm}}
    }
        \shortstack[c]{(a)\\Source\\\phantom{registered}}
        &
        \shortstack[c]{(b)\\Target\\unregistered}
        &
        \shortstack[c]{(c)\\Target\\registered}
        &
        \shortstack[c]{(d)\\Numerical~\cite{laga2017numerical}\\ registered}
        &
        \shortstack[c]{(e)\\Numerical~\cite{laga2017numerical}\\  unregistered}
        &
        \shortstack[c]{(f)\\Ours\\registered}
        &
        \shortstack[c]{(g)\\Ours\\unregistered}
    \end{tabular}
\end{adjustbox}%
\hspace*{2pt}

    \caption{Comparison of computational efficiency and inversion quality of mean 4D surface obtained using the numerical SRNF~\cite{laga2017numerical,nizamani2025DSNS} and the proposed \MethodName. The bottom row, which highlights a 3D instance from each 4D sequence, shows that the proposed \MethodName~produces more natural mean shapes while achieving a $180\times$ speedup; see the  Supplementary Material for additional results.}
    \label{fig:spatiotemporal_geodesics_neural_srnf_linear_comparison}
\end{figure*}

\begin{figure*}
    \centering
    \includegraphics[width=0.98\linewidth,trim={0cm 0cm 0cm 0cm},clip ]{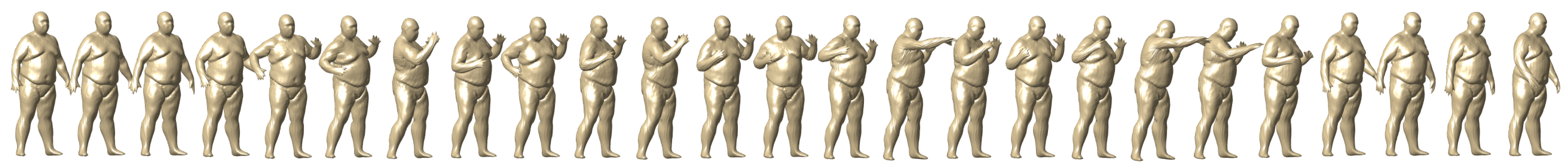}\\
    \includegraphics[width=0.98\linewidth,trim={0cm 0cm 0cm 0cm},clip ]{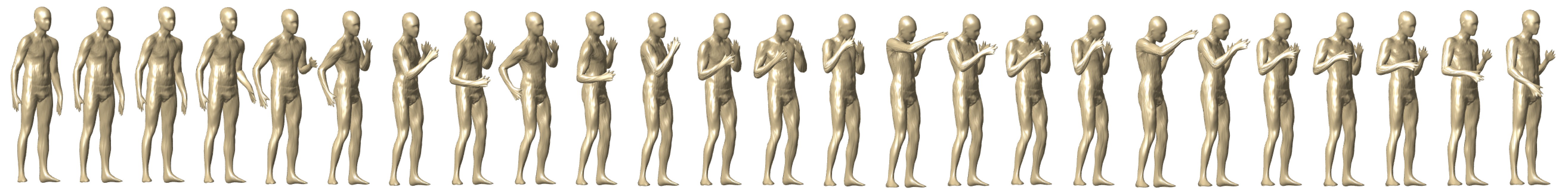}\\
    \includegraphics[width=0.98\linewidth,trim={0cm 0cm 0cm 0cm},clip ]{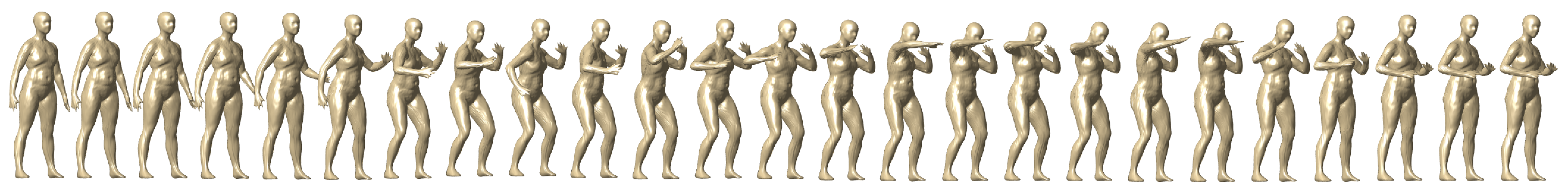}\\
    \includegraphics[width=0.98\linewidth,trim={0cm 0cm 0cm 0cm},clip ]{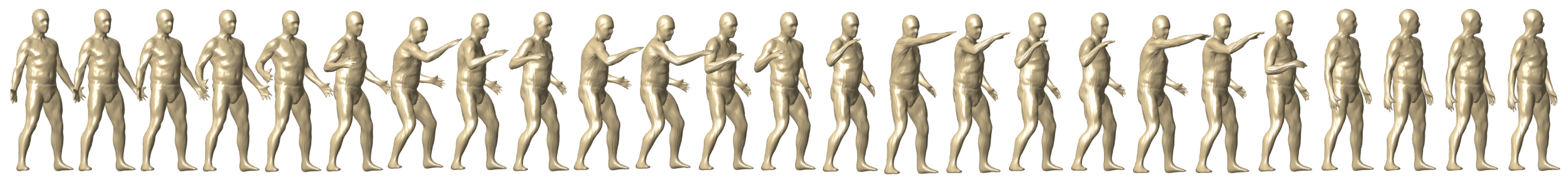}\\
    
    \includegraphics[width=0.98\linewidth,trim={0cm 0cm 0cm 0cm},clip ]{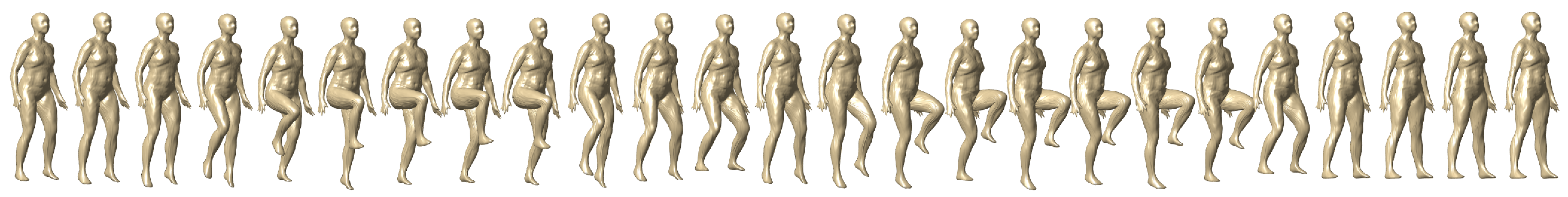}\\
    \includegraphics[width=0.98\linewidth,trim={0cm 0cm 0cm 0cm},clip ]{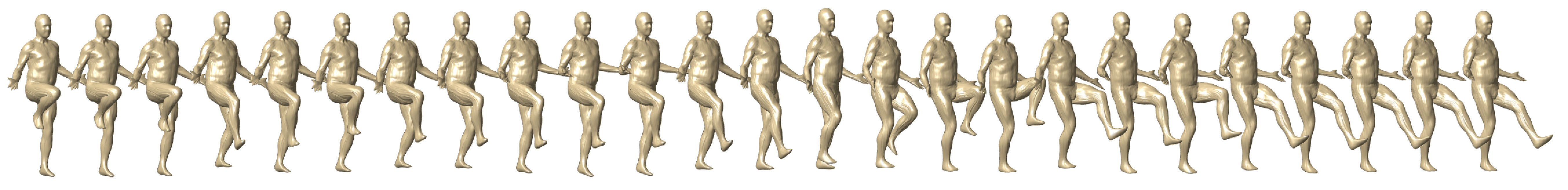}\\

    \rule{\linewidth}{0.5pt}\\

    \includegraphics[width=0.98\linewidth,trim={0cm 0cm 0cm 0cm},clip ]{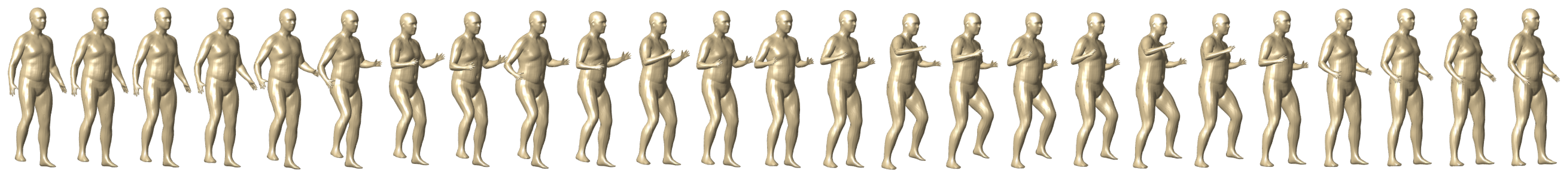}\\

    \textbf{(a)} Mean shape computed in the SRNF space and inverted using numerical SRNF~\cite{laga2017numerical} ($45$ min).\\    

    \includegraphics[width=0.98\linewidth,trim={0cm 0cm 0cm 0cm},clip ]{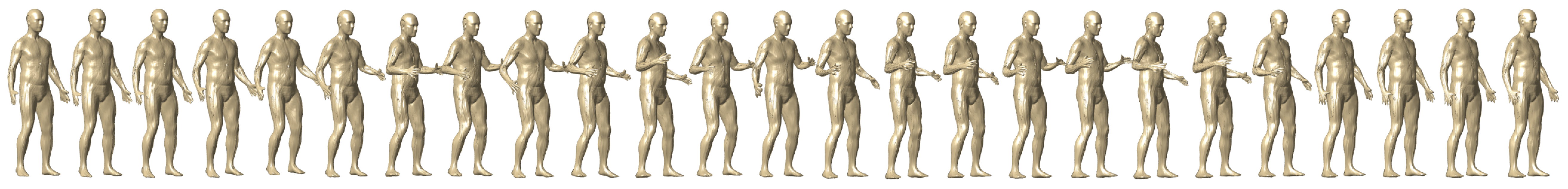}\\
    \textbf{(b)} Mean shape computed in the SRNF space and inverted using the proposed \MethodName~($15$ sec).\\

    \caption{Example of the mean 4D surface computed on six 4D surfaces from the DFAUST dataset after temporal registration. We show the mean computed using \textbf{(a)} numerical SRNF~\cite{laga2017numerical} and \textbf{(b)} the proposed \MethodName~method. Here, the first four rows perform the same punching action and the following two perform a one-leg jump action. Note that the proposed \MethodName~is $180$ times faster than the numerical method ($15$ s against $45$ min).}
    % with a $50$ times faster speedup; see the  Supplementary Material for additional results.}
    \label{fig:spatiotemporal_4D_mean}
\end{figure*}

\begin{figure}
    \centering

    \setlength{\tabcolsep}{1pt} 
    \renewcommand{\arraystretch}{0.95}
    
    \begin{tabular}{
        @{}
        >{\centering\arraybackslash}m{0.025\linewidth}
        @{\hspace{0pt}}
        >{\raggedright\arraybackslash}m{0.975\linewidth}
        @{}
    }

        \rotatebox[origin=c]{90}{ \scriptsize $+2.5\alpha$}&
        \includegraphics[width=\linewidth,trim={0cm 0cm 0cm 0cm},clip ]{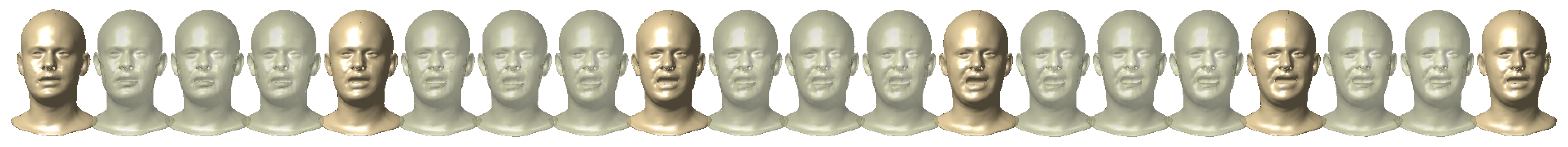}\\

        \rotatebox[origin=c]{90}{\scriptsize $+1.5\alpha$}&\includegraphics[width=\linewidth,trim={0cm 0cm 0cm 0cm},clip ]{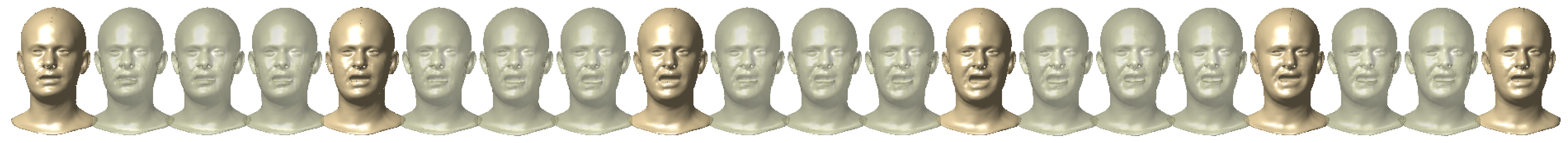}\\

        \rotatebox[origin=c]{90}{\scriptsize Mean}&\includegraphics[width=\linewidth,trim={0cm 0cm 0cm 0cm},clip ]{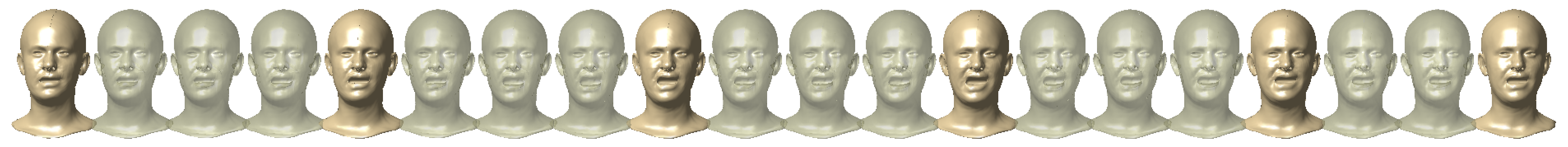}\\

        \rotatebox[origin=c]{90}{\scriptsize $-1.5\alpha$}&\includegraphics[width=\linewidth,trim={0cm 0cm 0cm 0cm},clip ]{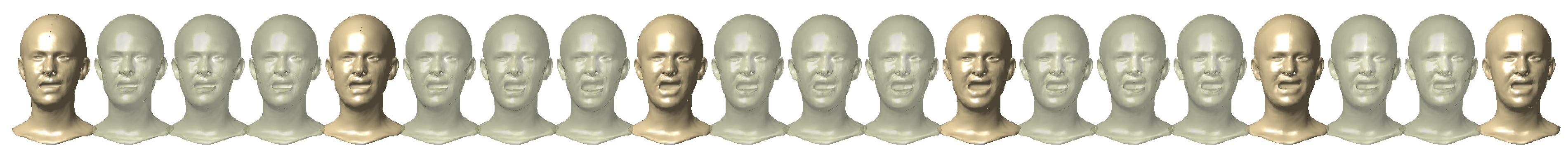}\\

        \rotatebox[origin=c]{90}{\scriptsize $-2.5\alpha$}&\includegraphics[width=\linewidth,trim={0cm 0cm 0cm 0cm},clip ]{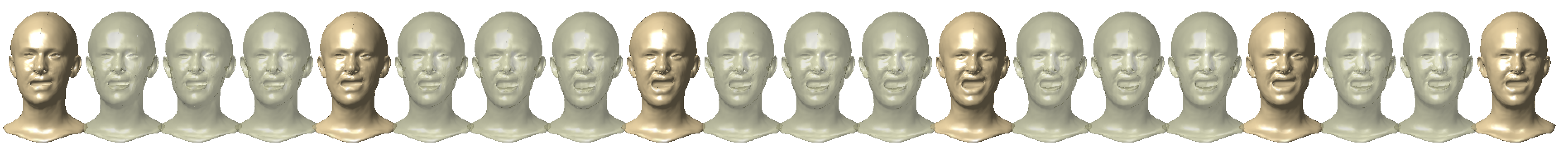}\\

    \end{tabular}
    
        \vspace{1.5pt}
    
        \rule{\linewidth}{1pt}
    
        \vspace{1.5pt}

    \setlength{\tabcolsep}{1pt} 
    \renewcommand{\arraystretch}{0.95}
    
    \begin{tabular}{
        @{}
        >{\centering\arraybackslash}m{0.025\linewidth}
        @{\hspace{0pt}}
        >{\raggedright\arraybackslash}m{0.975\linewidth}
        @{}
    }
        
        \rotatebox[origin=c]{90}{\scriptsize $+2.5\alpha$} & \includegraphics[width=\linewidth,trim={0cm 0cm 0cm 0cm},clip ]{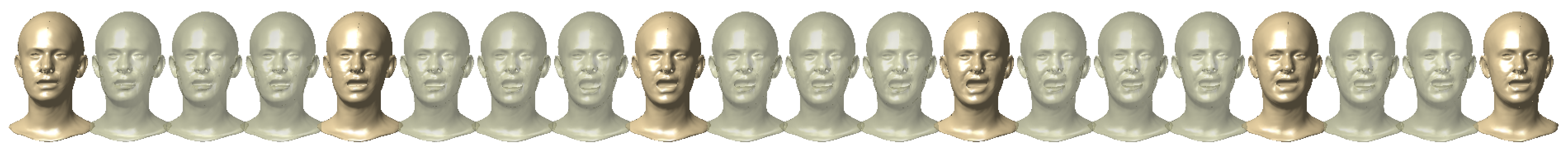}\\
        
        \rotatebox[origin=c]{90}{\scriptsize $+1.5\alpha$} & \includegraphics[width=\linewidth,trim={0cm 0cm 0cm 0cm},clip ]{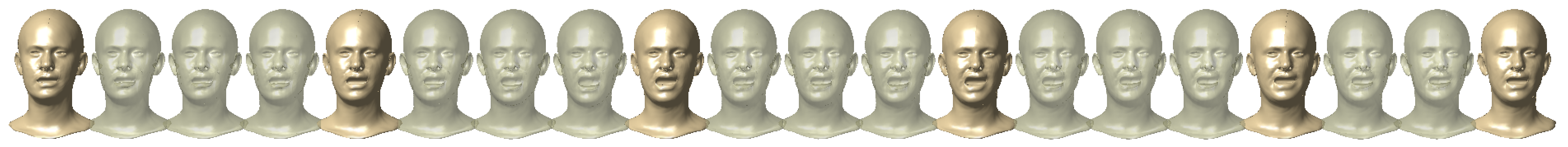}\\
        
        \rotatebox[origin=c]{90}{\scriptsize Mean} & \includegraphics[width=\linewidth,trim={0cm 0cm 0cm 0cm},clip ]{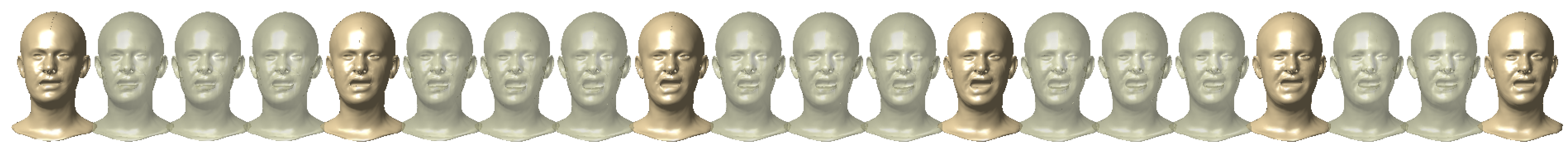}\\
        
        \rotatebox[origin=c]{90}{\scriptsize $-1.5\alpha$} & \includegraphics[width=\linewidth,trim={0cm 0cm 0cm 0cm},clip ]{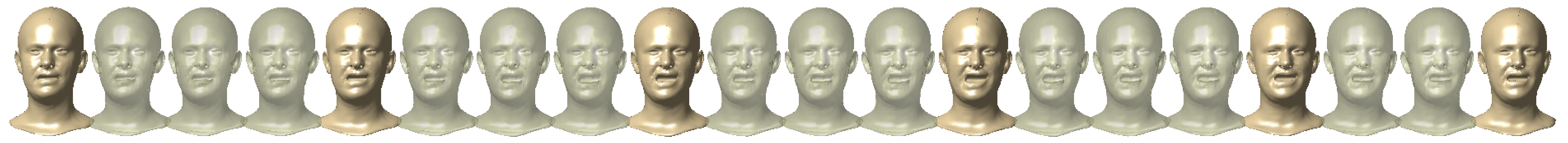}\\
        
        \rotatebox[origin=c]{90}{\scriptsize $-2.5\alpha$} & \includegraphics[width=\linewidth,trim={0cm 0cm 0cm 0cm},clip ]{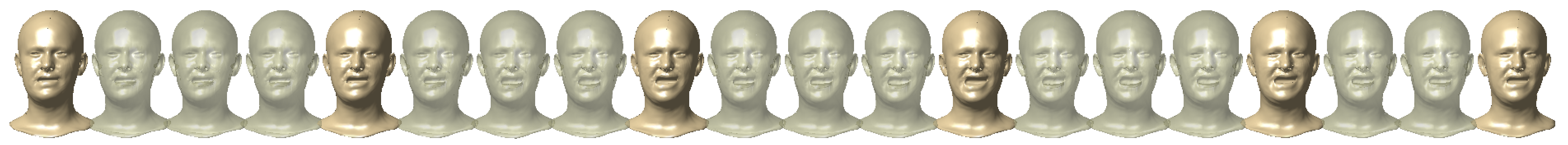}\\

    \end{tabular}
    
    \caption{Example of the two principal modes of variations computed on 4D faces in the SRNF space and inverted using the proposed \MethodName.}
    % with a $50$ times faster speedup; see the  Supplementary Material for additional results.}
    \label{fig:modes_of_variations_4D}
\end{figure}

% \begin{figure*}
%     \centering
%     \includegraphics[width=\linewidth,trim={0cm 0cm 0cm 0cm},clip ]{figures/deformation_transfer_temporal_2/Deformation_DFAUST_Visualizationsource.png}\\
%     \textbf{(a)} Source\\
%     \includegraphics[width=\linewidth,trim={0cm 0cm 0cm 0cm},clip ]{figures/deformation_transfer_temporal_2/Deformation_DFAUST_Visualizationtarget_linearly_transfered.png}\\
%     (b) Deformation transferred to the target using a linear approach.\\
%     \includegraphics[width=\linewidth,trim={0cm 0cm 0cm 0cm},clip ]{figures/deformation_transfer_temporal_2/Deformation_DFAUST_Visualizationtarget_srnf_transfered_numerical.png}\\
%     (c) Deformation transferred to the target using numerical SRNF~\cite{laga2017numerical}.\\
%     \includegraphics[width=\linewidth,trim={0cm 0cm 0cm 0cm},clip ]{figures/deformation_transfer_temporal_2/Deformation_DFAUST_Visualizationtarget_srnf_transfered.png}\\
%     (d) Deformation transferred to the target using the proposed \MethodName.

%     \caption{Example of the deformation transfer for 4D surfaces from the animal dataset.}
%     % with a $50$ times faster speedup; see the  Supplementary Material for additional results.}
%     \label{fig:deformation_trasnfer_4D}
% \end{figure*}

We evaluate quantitatively and qualitatively the quality of the geodesics and the computation time of the proposed method against state-of-the-art methods. 

Table~\ref{tab:pairwise_geodesic} reports the pairwise geodesic error. For each pair of source and target 3D shapes, we first compute a geodesic path using LIMP~\cite{cosmo2020LIMP}, NeuroMorph\cite{eisenberger2021neuromorph}, SAE\cite{lemeunier2022SAE}, and the proposed \MethodName. We then take each of the 3D shapes along the geodesic, sample $1000$ points, and compute the pairwise geodesic distances (\ie distances along the shape surface) between these points. We then define the error as the $\mathbb{L}^1$ distance between the pairwise geodesics of the source surface and the intermediate surfaces along the geodesic path. A lower error indicates a better geodesic path. As shown in the table, our method achieves lower mean, median, and standard deviation errors compared with the state-of-the-art methods. For a fair comparison, we evaluate each method on the datasets supported by the authors' released code.  LIMP~\cite{cosmo2020LIMP} and NeuroMorph~\cite{eisenberger2021neuromorph} implementations provide training code for FAUST~\cite{bogo2014faust}, while SAE~\cite{lemeunier2022SAE} provides code for DFAUST~\cite{dfaust:CVPR:2017}. Therefore, we compare each method only on the supported datasets.

Table~\ref{tab:geodesic_errors} reports the geodesic error using the evaluation metric proposed in LIMP~\cite{cosmo2020LIMP} on $100$ pairs from the FAUST~\cite{bogo2014faust}, DFAUST~\cite{dfaust:CVPR:2017}, and COMA~\cite{COMA:ECCV18} datasets. Since LIMP~\cite{cosmo2020LIMP} provides evaluation code only for FAUST~\cite{bogo2014faust}, we use the geodesic error results reported in its paper (Table~1) for this comparison. We follow a similar protocol to LIMP~\cite{cosmo2020LIMP}; we compute geodesic error over both training and test samples, allowing us to evaluate the geodesic error on $100$ pairs. Note that, during training, LIMP~\cite{cosmo2020LIMP} directly optimizes  this geodesic error. Although our method is not optimized on this geodesic error, it is able to achieve a lower error on FAUST~\cite{bogo2014faust} and COMA~\cite{COMA:ECCV18} than  LIMP~\cite{cosmo2020LIMP}.

Figure~\ref{fig:animal_geodesic_with_correspondence} shows an example of a geodesic path between two animals computed with our method after registration. Originally, the 3D shapes come with ground-truth registration. Thus, we first randomly reparameterize the lion $\surfacetwo$ to simulate the case where they are not in spatial correspondence; see Figure~\ref{fig:animal_geodesic_with_correspondence}-\textbf{(a)}. To bring the surfaces in correspondence, we use the DSNS method~\cite{nizamani2025DSNS} to estimate the spatial reparameterization of $\surfacetwo$ and align it to $\surfaceone$ using the $\ltwo$ metric in the SRNF space. Figure~\ref{fig:animal_geodesic_with_correspondence}-\textbf{(b)} shows the surfaces after registration. We then compute the geodesic between the two registered surfaces in the SRNF space and reconstruct the surfaces using the proposed \MethodName.  Figure~\ref{fig:animal_geodesic_with_correspondence}-\textbf{(c)} shows the geodesic path from the top-view.

Figure~\ref{fig:animal_1_geodesics_main} compares the quality of the geodesic shown in Figure~\ref{fig:animal_geodesic_with_correspondence} (shown from a front view) against linear interpolation in the original space $ \surfaces$ of surfaces (\ie $(1-t)\surfaceone + t\surfacetwo, t \in [0,1]$) and the geodesic in $\srnfs$ but computed using the numerical SRNF inversion procedure of~\cite{laga2017numerical}. From this experiment, we can clearly see that the linear interpolation results have unnatural shrinkage, especially in parts such as the head, which undergoes substantial bending. This is expected since linear interpolation assumes that the surfaces are elements of a Euclidean shape space equipped with the $\ltwo$ metric, which is not the case for 3D shapes. The geodesic computed using the numerical method of Laga \etal~\cite{laga2017numerical} shows improvement in quality but still exhibits unnatural shrinkage along the geodesic path. On the other hand, the geodesic computed using the proposed \MethodName~is natural: observe, for example, how the head of the animal in Figure~\ref{fig:animal_1_geodesics_main}-\textbf{(c)} bends naturally, instead of shrinking as in Figure~\ref{fig:animal_1_geodesics_main}-\textbf{(c)}. Figure~\ref{fig:animal_geodesics_2_suppl} shows an additional example rendered from the top-view. Here we can see that the tail movement looks more natural between the two foxes in Figure~\ref{fig:animal_geodesics_2_suppl}-\textbf{(b)}, whereas the numerical method~\cite{laga2017numerical} in Figure~\ref{fig:animal_geodesics_2_suppl}-\textbf{(a)} exhibits unnatural shrinkage.

Figures~\ref{fig:human_geodesics_implicit_comparison_1} and~\ref{fig:limp_1_geodesics_main} compare the quality of the geodesics between 3D humans computed using linear interpolation, numerical SRNF inversion~\cite{laga2017numerical}, LIMP~\cite{cosmo2020LIMP}, NeuroMorph~\cite{eisenberger2021neuromorph}, 4Deform~\cite{sang20254deform},  the implicit neural surface approach of Sang \etal~\cite{sang2025implicit},  and with the proposed \MethodName. In this experiment, we considered the geodesic failure case reported in~\cite{sang2025implicit}. We can clearly see that, in the case of our method (Figures~\ref{fig:human_geodesics_implicit_comparison_1}-\textbf{(e)} and~\ref{fig:limp_1_geodesics_main}-\textbf{(c)}), the arm bends and stretches naturally, whereas previous methods (Figures~\ref{fig:human_geodesics_implicit_comparison_1}-\textbf{(a)}, \textbf{(b)}, \textbf{(c)}, \textbf{(d)},  and Figures~\ref{fig:limp_1_geodesics_main}-\textbf{(a)}, \textbf{(b)}) fail to recover plausible intermediate geodesics. The Supplementary Material provides additional comparisons. 

In terms of computation time, our approach computes the geodesic of Figure~\ref{fig:animal_1_geodesics_main} in $2$ s, whereas the numerical method~\cite{laga2017numerical} takes $118$ s. As a result, our method achieves an approximately $50$-fold speedup. In the case of Figure~\ref{fig:human_geodesics_implicit_comparison_1} and~\ref{fig:limp_1_geodesics_main}, our approach takes $3$ s compared to $7$ min for  4Deform~\cite{sang20254deform}, $20$ min for the neural implicit surface~\cite{sang2025implicit}, $1$ sec for the LIMP~\cite{cosmo2020LIMP}, and $1$ sec for NeuroMorph~\cite{eisenberger2021neuromorph}. LIMP~\cite{cosmo2020LIMP} and NeuroMorph~\cite{eisenberger2021neuromorph}, however, are discrete approaches and reconstruct meshes on a fixed set of vertices, limiting reconstruction quality, whereas our method learns a continuous and resolution-agnostic representation that results in higher quality meshes and does not require mesh connectivity.  Note that implicit methods~\cite{sang2025implicit,sang20254deform} require training for each pair of shapes, whereas our approach is trained on the collection of 3D shapes and only requires inference to compute geodesics. 

We compare our method qualitatively and quantitatively with the most closely related approaches of 3D geodesics. However, we could not include \cite{Yang2023GeoLatent} in this comparison as no official implementation, pretrained model, or generated outputs were publicly available at the time of submission\footnote{We contacted the authors to request access, but did not receive a response. We tried independent reimplementation, but due to a lack of training and inference protocols, we were unable to reproduce visual qualitative comparisons.}.

\subsection{Deformation transfer}
\label{sec:results_deformation_transfer}

Figure~\ref{fig:deformation_transfer_three_examples_with_extrapolation} shows examples of deformation transfer across human, animal, and facial shapes obtained using the approach presented in Section~\ref{sec:deformation_transfer}. The source deformations, \ie the surface $\surfaceone$ that deforms into the surface $h_1$, are shown in Figure~\ref{fig:deformation_transfer_three_examples_with_extrapolation}-(a). Figures~\ref{fig:deformation_transfer_three_examples_with_extrapolation}-(b) and (c) show the  transfer of these deformations to another surface  using  linear extrapolation in the SRNF space, starting from $\srnfmap(\surfacetwo)$ to $\alpha_{\srnf}(\tau) = \srnfmap(\surfacetwo) + \tau \srnfmap(v)$.  With a slight abuse of notation, let $\srnfmap(v) = \srnfmap(h_1)- \srnfmap(\surfaceone)$. We then   map them to the space of surfaces $\surfaces$ using the numerical SRNF inversion method~\cite{laga2017numerical} (Figure~\ref{fig:deformation_transfer_three_examples_with_extrapolation}-(b)) and the proposed \MethodName~(Figure~\ref{fig:deformation_transfer_three_examples_with_extrapolation}-(c)). 
In both cases, we show results for $\tau \in [0, 1.5]$, at intervals of $0.25$.  $\tau = 1$ corresponds to $h_2 = \alpha(1)$ should have the same pose as $h_1$. The cases of $\tau>1$ correspond to extrapolation beyond the range of the source deformation. As one can see in Figure~\ref{fig:deformation_transfer_three_examples_with_extrapolation}-(b), deformation transfer using numerical SRNF~\cite{laga2017numerical} results in unnatural poses where the limbs unnaturally shrink. This shrinkage becomes significant as we extrapolate further away from the target pose.  On the other hand, our method (Figure~\ref{fig:deformation_transfer_three_examples_with_extrapolation}-(c))  obtains  natural deformation transfer. As one can see,  $h_2 = \alpha(1) $ corresponds exactly to the same pose as $h_1$. We can also observe that extrapolating beyond $\tau = 1$, using our \MethodName\ framework, still results in natural poses.

\subsection{Summary statistics and 3D shape generation}
\label{sec:results_statistics}

We evaluate the performance of the proposed framework in computing summary statistics and generating novel 3D shapes. Given a set of 3D shapes, we first  normalize them for translation and scale, and convert each to its SRNF representation. We then compute the statistical summaries in the SRNF space following the algorithm presented in Section~\ref{sec:summary_statistics}, and map them back to the original space $\surfaces$ for visualization, using our proposed method. We also use the computed mean shape and modes of variation to generate random surfaces in the SRNF space, using Eqn.~\eqref{eq:sampling}, and invert them to the shape space $\surfaces$ using our SRNF inversion approach. Figure~\ref{fig:animal_human_summary_statistics} shows the statistical mean and the first three modes of variation computed on the animal and human dataset. Figure~\ref{fig:animal_shape_generation} shows some 3D animals generated using the proposed \MethodName. Figure~\ref{fig:human_shape_generation_comparison} compares the quality of 3D human models generated using  numerical SRNF~\cite{laga2017numerical} (Figure~\ref{fig:human_shape_generation_comparison}-\textbf{(a)}) with those generated using our method (Figure~\ref{fig:human_shape_generation_comparison}-\textbf{(b)}). We can clearly see in the close-up views that the random shapes computed with our method capture all the characteristic features of the dataset. On the other hand, those computed using the numerical SRNF~\cite{laga2017numerical} have unnatural artifacts such as missing and shrunk limbs.

\subsection{Application to 4D surfaces}
\label{sec:application4Dsurfaces}

We now demonstrate the applicability of the proposed \MethodName~to 4D surfaces. A 4D (or 3D$+$time) surface is a 3D surface that evolves over time, \eg a moving 3D body shape, a facial surface that performs some expressions, or a surface of an anatomical organ that grows over time (due to the aging process).  
Following~\cite{nizamani2025DSNS}, we represent each 3D surface along a 4D sequence using its SRNF. Thus, a 4D surface becomes curves in the SRNF space. The problem of temporal alignment and geodesics and summary statistics computation between 4D surfaces becomes that of temporal alignment and geodesics and summary statistics computation  between curves in the SRNF space, which has an $\ltwo$ structure, and then mapping the results back to the space of surfaces using SRNF inversion.

%Figure~\ref{fig:human_1_spatiotemporal} shows an example of temporal alignment of 4D humans performing the jumping action from the DFAUST dataset at different speeds. In this figure, \textbf{(a)} is the source 4D surface, \textbf{(b)} is the target 4D surface, and \textbf{(c)} is the target 4D surface temporally aligned to the source 4D surface using DSNS~\cite{nizamani2025DSNS} method.}

Figure~\ref{fig:spatiotemporal_geodesics_neural_srnf} shows an example of joint temporal alignment (using DSNS~\cite{nizamani2025DSNS}) and geodesics computation  between two 4D  surfaces performing a punching action. We compute the geodesic between the top row (source 4D surface) and the bottom row (target 4D surface) in the SRNF space and invert the geodesic path to the original surface space using the proposed \MethodName. Compared to the numerical SRNF~\cite{laga2017numerical} used in DSNS~\cite{nizamani2025DSNS}, which takes around $45$ min to invert one 4D surface, the proposed  \MethodName~ is $180$ times faster (around $15$ s per 4D surface)  and more accurate.

Figure~\ref{fig:spatiotemporal_geodesics_neural_srnf_linear_comparison}  compares the computation time and accuracy of the mean 4D surfaces obtained using the numerical SRNF~\cite{laga2017numerical,nizamani2025DSNS} and the proposed \MethodName~method. Figures~\ref{fig:spatiotemporal_geodesics_neural_srnf_linear_comparison}-\textbf{(a)} and  \textbf{(b)} show two 4D surfaces that perform a jumping action before temporal alignment. We align the source 4D surface (Figure~\ref{fig:spatiotemporal_geodesics_neural_srnf_linear_comparison}-\textbf{(a)}) to the target surface in Figure~\ref{fig:spatiotemporal_geodesics_neural_srnf_linear_comparison}-\textbf{(c)}. We then compute the mean 4D surfaces in the SRNF space and invert them back to the surface space using numerical SRNF~\cite{laga2017numerical} and \MethodName. Figures~\ref{fig:spatiotemporal_geodesics_neural_srnf_linear_comparison}-\textbf{(d)} and \textbf{(e)} show the mean 4D surfaces obtained   using numerical SRNF  after and before  temporal registration, respectively. Note that it takes $45$ minutes to compute the inversion using numerical SRNF~\cite{laga2017numerical,nizamani2025DSNS}. On the other hand, Figures~\ref{fig:spatiotemporal_geodesics_neural_srnf_linear_comparison}-\textbf{(f)} and \textbf{(g)} show the mean 4D surface obtained using the proposed \MethodName~after   and before temporal registration, respectively. Notice that \MethodName~only takes $15$ s to generate the entire 4D sequence while being more accurate than the numerical method. Also, we show a close-up view of a selected 3D instance from the 4D sequences at the bottom. As one can see, the proposed \MethodName~is able to generate a plausible mean shape compared to the numerical SRNF~\cite{laga2017numerical,nizamani2025DSNS}.

Figure~\ref{fig:spatiotemporal_4D_mean} shows an example of temporal co-registration of multiple 4D surfaces, from the DFAUST dataset after temporal alignment, and their corresponding mean 4D surface. %We take the source 4D surface (top row) and temporally align it to the following five target surfaces (from second to sixth row). 
Here, the first four rows are performing a punching action and the last two rows are performing a one-leg jump action. We temporally co-aligned all these 4D surfaces using DSNS~\cite{nizamani2025DSNS} approach and show in Figure~\ref{fig:spatiotemporal_4D_mean}-\textbf{(a)} the 4D mean surface obtained using the numerical SRNF~\cite{laga2017numerical,nizamani2025DSNS} approach, which takes $45$ minutes and results in arm shrinkage.  In Figure~\ref{fig:spatiotemporal_4D_mean}-\textbf{(b)}, we show the 4D mean surface obtained using  the proposed \MethodName. We can see that the latter, which is generated in less than $15$ s,  looks more natural.

Figure~\ref{fig:modes_of_variations_4D} shows an example of the mean and the first two principal modes of variation of  4D faces from COMA dataset~\cite{COMA:ECCV18}. As one can see, the 4D mean is  able to capture the main facial features.

% \subsection{Margins and page numbering}

% All printed material, including text, illustrations, and charts,
% must be kept within a print area %6.31 inches (16.03 cm) 
% 7 inches (17.7 cm) wide by
% %9.10 inches (23.13 cm) 
% 9.44 inches (24 cm) high. Do not write or print anything
% outside the print area. Number your pages on odd sites right
% above, on even sites left above, no page number on the first site.
% Do not use page numbering within the final version of your paper.

%------------------------------------------------------------------------

\section{Conclusion}
\label{sec:conclusion}

We propose \MethodName, a novel continuous surface-based neural representation that provides a practical solution to the ill-posed SRNF inversion problem. Our solution is computationally very efficient while achieving significantly better accuracy than traditional methods. We further demonstrate how core statistical analysis tasks can be performed in the SRNF space, where the complex elastic metric reduces to an $\ltwo$ metric, facilitating downstream analysis tasks such as geodesic computation, deformation transfer, summary statistics, and shape generation. Our approach achieves state-of-the-art results for 3D shape analysis across human body, face, and animal datasets. Importantly, it is up to $100$ times faster than existing methods while achieving improved accuracy. Although efficient,  our approach requires registered 3D surfaces, needs to be trained separately for each shape class, and is limited to genus-0 surfaces.

\section*{Acknowledegment}
This work is supported by the Australian Research Council (ARC) Discovery Projects no. DP210101682, DP220102197,  and DP260101891, and ARC Future Fellowship FT250100448.

\clearpage

\bibliographystyle{abbrv-doi-hyperref}

\bibliography{main}

\end{document}